\documentclass{article} %
\usepackage{iclr2026_conference,times}

\usepackage{amsmath,amsfonts,bm}

\def\eqref#1{equation~\ref{#1}}

\def\1{\bm{1}}

\DeclareMathAlphabet{\mathsfit}{\encodingdefault}{\sfdefault}{m}{sl}
\SetMathAlphabet{\mathsfit}{bold}{\encodingdefault}{\sfdefault}{bx}{n}

\usepackage[colorlinks=true, linkcolor=blue, urlcolor=blue, citecolor=blue]{hyperref}
\AtBeginDocument{%
}
\usepackage{url}
\usepackage{xspace}
\usepackage{tcolorbox}
\usepackage{listings}
\usepackage{cuted}
\usepackage{capt-of}
\usepackage{float}
\usepackage{xstring}
\usepackage{pgfplots}
\usepackage{xcolor}
\usepackage{subcaption}
\usepackage{graphicx}
\usepackage{CJKutf8}
\usepackage{tabularx}
\usepackage{chngcntr}

\makeatletter
\newcommand\smallfootnotesize{\@setfontsize\smallfootnotesize{8}{9}}

\makeatother

\DeclareUnicodeCharacter{FE0F}{}
\usepackage{hwemoji}

\title{Constantly Improving Image Models Need \\ Constantly Improving Benchmarks}

\author{
Jiaxin Ge\textsuperscript{1}\thanks{Equal contribution. Correspondence to \texttt{gejiaxin@berkeley.edu}, \texttt{graceluo@berkeley.edu}}
\quad
Grace Luo\textsuperscript{1}\footnotemark[1]
\\[2pt]
\textbf{Heekyung Lee}\textsuperscript{1}
\quad
\textbf{Nishant Malpani}\textsuperscript{1}
\quad
\textbf{Long Lian}\textsuperscript{1}
\quad
\textbf{XuDong Wang}\textsuperscript{1}
\\[2pt]
\textbf{Aleksander Holynski}\textsuperscript{1}
\quad
\textbf{Trevor Darrell}\textsuperscript{1}
\quad
\textbf{Sewon Min}\textsuperscript{1}
\quad
\textbf{David M.~Chan}\textsuperscript{1}
\\[6pt]
\textsuperscript{1}UC Berkeley
}

\usepackage{comicneue}

\newcommand{\datasetnameplain}{ECHO}
\newcommand{\datasetname}{\texorpdfstring{{\comicneue{\datasetnameplain}}}{\datasetnameplain}}

\newcommand{\oaiimagegen}{4o Image Gen}
\newcommand{\twitter}{Twitter/X}

\pgfplotsset{compat=1.18}

\iclrfinalcopy %
\usepackage{todonotes}
\usepackage{wrapfig}
\usepackage{subcaption}
\usepackage{booktabs}
\usepackage{enumitem}
\usepackage[subtle]{savetrees}

\renewcommand{\paragraph}[1]{\vspace{0.5em}\noindent\textbf{#1}$\:\:$}
\begin{document}

\maketitle

\begin{abstract}
Recent advances in image generation, often driven by proprietary systems like GPT-4o Image Gen, regularly introduce new capabilities that reshape how users interact with these models. Existing benchmarks often lag behind and fail to capture these emerging use cases, leaving a gap between community perceptions of progress and formal evaluation. To address this, we present \datasetname{}, a framework for constructing benchmarks directly from real-world evidence of model use: social media posts that showcase novel prompts and qualitative user judgments. Applying this framework to GPT-4o Image Gen, we construct a dataset of over 31,000 prompts curated from such posts. Our analysis shows that \datasetname\ (1) discovers creative and complex tasks absent from existing benchmarks, such as re-rendering product labels across languages or generating receipts with specified totals, (2) more clearly distinguishes state-of-the-art models from alternatives, and
(3) surfaces community feedback that we use to inform the design of metrics for model quality (e.g., measuring observed shifts in color, identity, and structure). Our website is at \url{https://echo-bench.github.io}.
\end{abstract}

\section{Introduction}
\label{sec:intro}

When new generative image models are released, users often find new and unanticipated capabilities not captured by existing benchmarks. These capabilities are discussed on social media, where users document their interactions with new models and qualitatively discuss their performance. The release of GPT-4o Image Gen~\citep{openai2025} exemplified this behavior with the introduction of ``Ghiblification,'' the style-transfer task of turning a natural image into a cartoon version emulating a particular animated studio. This new ``task'' was not only shared widely on social media, but used as a personal measure of model quality by many members the online community. As of today, explicit benchmarks have now been developed for this task \citep{jiang2025balanced}, but the benchmarks that we traditionally use to evaluate models do not have the capability to evolve \textit{with} community feedback, and instead, must \textit{react} to changes in a delayed cycle. 

\begin{figure*}
    \centering
    \captionsetup{type=figure}
    \includegraphics[width=\textwidth]{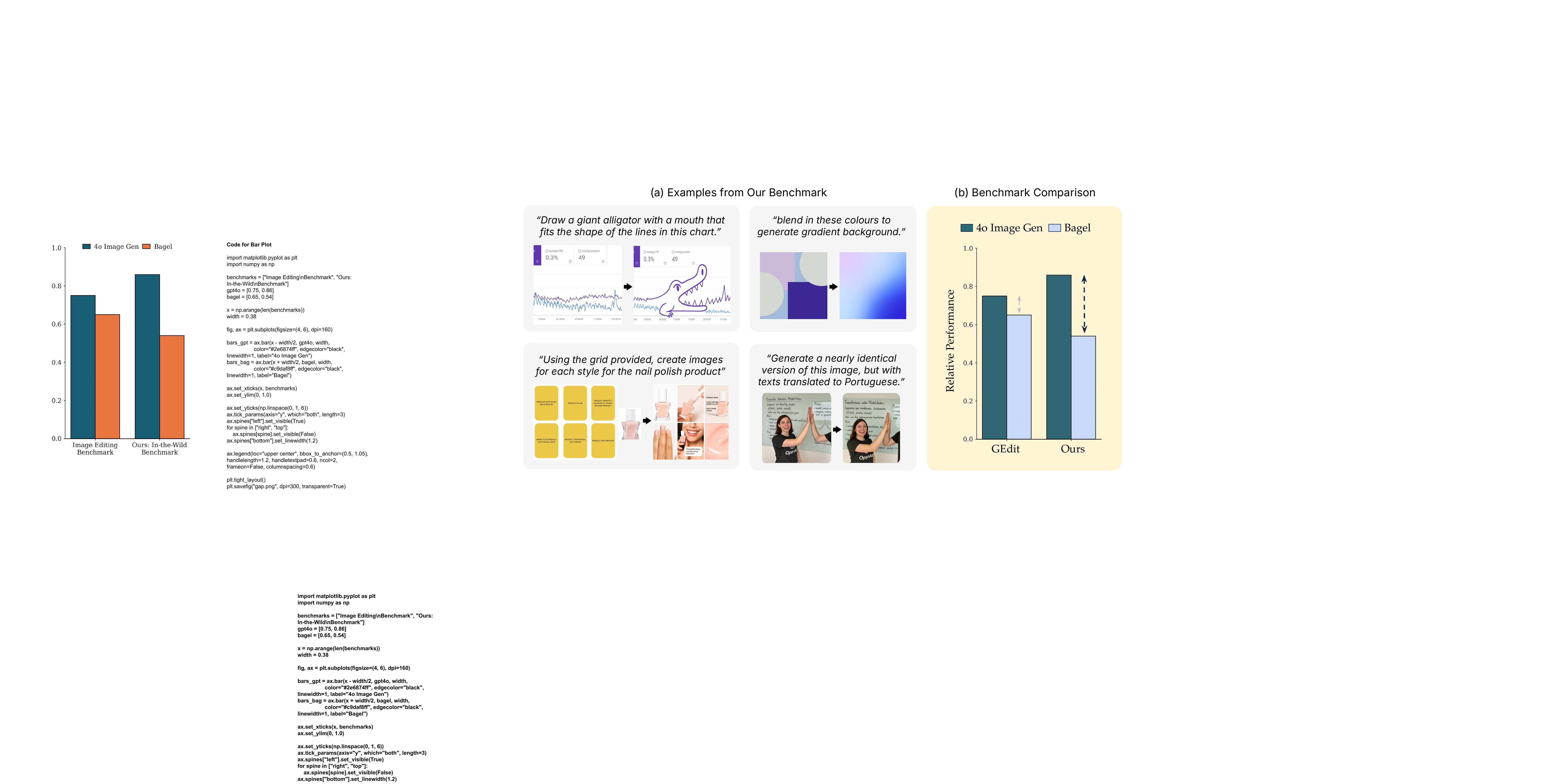}
    \captionof{figure}{\textbf{\datasetname{}} distills collective discussion about a new generative model into a structured benchmark. As a case study, we apply \datasetname{} to GPT-4o Image Gen~\citep{openai2025} on Twitter/X. Left: \datasetname{} \textit{automatically} surfaces highly diverse and novel tasks not covered in prior benchmarks.
    Right: Consequently, our image-to-image split shows a 3.2x larger relative performance gap compared to a prior image editing benchmark, GEdit~\citep{liu2025step1x}.
    }
    \label{fig:teaser}
\end{figure*}

Indeed, despite significant changes in what constitutes a ``good'' image generation model, current popular crowdsourced text-to-image benchmarks~\citep{wang2022diffusiondb, kirstain2023pick} are often still tailored towards older models such as Stable Diffusion~\citep{rombach2022high}, with extensive art-centric keyword lists that are not representative of now-feasible use cases. Popular image editing benchmarks ~\citep{Zhang2023MagicBrush, zhang2023hive, liu2025step1x} contain overly simple instructions. These instructions were challenging at their inception but do not actually require complex language understanding or reasoning. Furthermore, these tasks can already be solved by many models, new and old. This slow adaptation rate is reflected in model benchmark scores. As we see in \hyperref[fig:teaser]{\autoref*{fig:teaser}b}, human ratings indicate that 4o Image Gen is substantially better than the current best open-source unified model~\citep{deng2025emerging}, yet even when benchmarking on a recent image editing benchmark~\citep{liu2025step1x}, the gap appears less significant.

With the rapid releases of new image generation models, each revealing a range of new capabilities to be tested, it has become clear that we need more responsive mechanisms for adapting benchmarks to emergent user observations.
In this work we present \datasetname{}: \underline{E}xtracting \underline{C}ommunity \underline{H}atched  \underline{O}bservations, a \textit{re-usable framework} that converts community discussion on social media into a structured benchmark. Our proposed method bypasses the traditional ``observation to benchmark'' cycle, and provides us with a framework for \textit{automatically} converting real-world ideas and capabilities surfaced by users on social media directly to metrics that we can use to measure and improve SOTA models. 
The \datasetname{} framework operates by searching social media for mentions of a target model and automatically filtering for coherent image generation prompts specified via text and/or images, while extracting community insights and feedback on particular prompt capabilities. It is designed to address a number of common challenges associated with social media, including the tradeoff between post volume and relevance, the splitting of context across posts, and noisy formatting.

Using \datasetname{}, we are able to surface and formalize a number of qualitative observations related to the most recent image generation methods. By running our framework on the 4o Image Gen release, we introduce a new dataset containing more than 31,000 user-sourced prompts which: (1) surfaces creative and complex tasks absent from existing benchmarks, (2) is more diverse and more closely resembles natural user language (contains 2.3x more unique first bigrams and are 1.2x lower in LLM perplexity), (3) better separates state-of-the-art models from prior models, and (4) \textit{automatically} surfaces several new quantifiable indicators for image generation quality, including identity preservation and color shift, which we show can be operationalized into secondary evaluation metrics that could inform future model losses and development.

\section{Background \& Related Work}
\label{sec:inadequacy}

Model benchmarks often mirror the capabilities of the models themselves, and are designed by model developers in order to evaluate and understand how these models perform on tasks of interest. For example, traditional text-to-image benchmarks~\citep{huang2023t2i, ghosh2023geneval, lee2023holistic} and image-to-image benchmarks~\citep{brooks2023instructpix2pix, wang2023imagen, sheynin2024emu, hui2024hq, Zhang2023MagicBrush} are not collected in-the-wild. These benchmarks contain short and overly simple instructions such as ``A cat in front of a chair" or ``Add fireworks in the sky" that fail to reflect real user intent, but provide strong diagnostic signal for understanding simple generative understanding. 

On the other hand, community-driven benchmarks are often designed to collect real user prompts, and more closely mirror what a downstream user might desire from a model. For example, previous methods~\citep{wang2022diffusiondb, xu2023imagereward, kirstain2023pick} have collected real user prompts of Stable Diffusion models from an explicit interface~\citep{rombach2022high}. These benchmarks require a significant amount of human intervention to decide which user prompts to model, and how to model them. In addition, 
the model interface itself can lead to two further limitations: (i) prompt intent is bounded by the capabilities of the model itself, and (ii) prompt style is tailored towards the model rather than natural user language. For example, it has already been demonstrated that users will adjust their prompting behavior to account for limitations of the CLIP~\citep{radford2021learning} text encoder, which behaves more like a bag-of-words representation, where they use extensive sets of ``phrases rather than complete sentences''~\citep{comfyui_prompt_basic_2025}, with prompts like \textit{``colorful stars,galaxies,space,artstation.''} Unlike interface-collected datasets, our framework draws on prompts crafted for human audiences on social media, where the goal is to showcase creativity rather than to optimize around model quirks. While prompts inevitably reflect the capabilities of the current best models, our framework is re-runnable and can adapt as models and user behaviors evolve, reducing the risk of per-model biases.

GEdit~\citep{liu2025step1x} proposed scraping the internet for real image editing prompts. However, these prompts are limited by the imagination of the authors, leading to a restricted set of 11 specific single image editing tasks, such as changing the color or changing the background. Most closely related to our work, IntelligentBench~\citep{deng2025emerging} and KontextBench~\citep{batifol2025flux} were designed to highlight the capabilities of new models released by the same authors. However, details about their data source and creation method are largely unknown, and neither benchmark is publicly available.

Outside of image generation, Chatbot Arena~\citep{chiang2024chatbot} uses an online platform to collect use cases in the wild, incentivizing users to provide data by providing a free platform for interacting with the model. While such a process does collect real user prompts, unlike this approach, we investigate social media, which represents a notably different prompt distribution: since users are seeking reciprocal engagement, they are more incentivized to produce novel and creative examples, rather than tasks that are already well-within model capabilities.

\section{Crowdsourcing a Benchmark}
\label{sec:method}

\begin{figure}[t]
    \centering
    \includegraphics[width=\linewidth]{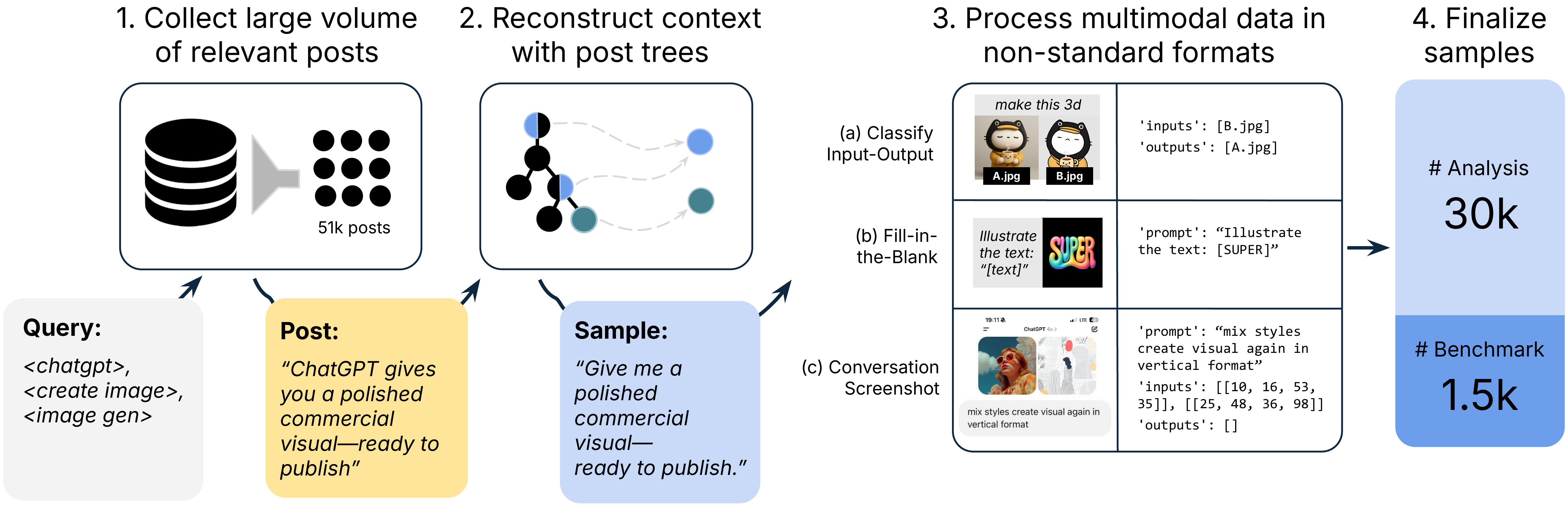}
    \caption{\textbf{\datasetname{} Framework.}
    \datasetname{} is motivated by several challenges inherent to social media. (1) 
    We start with broad queries followed by relevance filtering, since basic querying presents a volume-relevance tradeoff. (2) We then extract prompts from these posts, making sure to utilize the full post tree, as context can be spread across posts. (3) We then apply multimodal processing, since useful data also exists in non-standard formats. (4) Finally, we reserve the highest quality data for benchmarking, while the rest is used for analysis.
    }
    \label{fig:pipeline}
\end{figure}

Our primary goal is to distill collective discussion about a new generative model into a structured dataset.
Such discussion often involves users sharing interesting prompts and outputs, novel task ideas, or commentary on model behavior.
We aim to capture all of these cases, in a standardized format:
\[
\smallfootnotesize \verb|<input text, input image(s)*, output image, community feedback*>| \quad
\]
where * denotes optional fields; the full set of data we collect is given in \autoref{tab:data_fields_full}.
However, this objective poses several challenges:
\begin{itemize}[left=1em]
\item Collection: A large volume of relevant data is desired, which requires identifying the right platform and gathering the data.
\item Processing: A non-trivial amount of processing is required, e.g., the input prompt and images may be embedded in a single screenshot  or the prompt may not be written explicitly.
\item Filtering: Data quality varies widely, e.g., a user may provide more general commentary or exactly document their input prompt.
\end{itemize}

We propose a framework, \datasetname{}, that addresses these challenges, illustrated in~\autoref{fig:pipeline}. 
Our framework first collects relevant posts (\autoref{subsec:maximizing_volume}), converts posts into self-contained samples (\autoref{subsec:reconstructing_context}), and finally expands coverage via multimodal processing (\autoref{subsec:processing_multimodal}). 

\subsection{Identifying Relevant Posts}\label{subsec:maximizing_volume}
There is an inherent tradeoff between the volume of posts and their relevance. When querying with broader keywords, the average post relevance goes down, and with narrower ones, the available post pool is quickly exhausted. To address this, we implement a two-stage pipeline where we first query for a large volume of posts then use an LLM to filter irrelevant ones.

\textbf{Designing Keywords.}
First, our goal is to maximize the post pool. However, we found two issues: (1) LLM-based filtering is expensive, so the pool cannot be too large, and (2) there is a temporal shift in which keywords lead to relevant posts (e.g., in the initial two weeks of the 4o Image Gen release, generic terms like ``openai'' often retrieve relevant posts, but later on relevancy decreases). Therefore, we use two sets of keywords to query posts within vs. outside the first two weeks of the 4o Image Gen release (see~\autoref{fig:query_keywords}).

\textbf{Classifying Relevance.}
We then use an LLM to classify the post text on a 5-point relevance scale (see~\autoref{fig:0_coarse_filter}).
We initially collected 68k posts in total, of which 47\% passed our relevance filter as `very likely relevant'' or ``certainly relevant.''
Nearly half of collected posts pass this filter, amounting to 32k posts, indicating that our query design is fairly efficient and has a high yield rate.

\subsection{Reconstructing Context Across Posts} \label{subsec:reconstructing_context}
Posts can be context dependent. For example, a user may write \textit{``prompt below''} in the first post then include the actual prompt text in a reply.
We want self-contained \textit{samples}, characterized as: a unique prompt some user tried, community feedback towards that prompt and its resulting outputs, and a label for its quality. To achieve this, our framework 
attempts to collect as much of the reply tree as possible, then use this full context when processing posts into samples.

\textbf{Constructing Reply Trees.}
For each post obtained via keyword query, we extract the full reply tree, or URLs pointing to the parent post or child replies. We then recursively expand the dataset by querying these discovered posts and traversing their respective reply chains, introducing 19k new posts from the replies. This procedure enables our framework to discover relevant posts that may not otherwise appear with keyword-based queries.
After reply collection, each post contains ancestor chain $\mathcal{P}_{\uparrow} = \langle P_0, \dots, P_n \rangle$ and direct replies $\mathcal{C}_{\downarrow} = \{ C_0, \dots, C_m \}$.
We then search for the unique reply trees across all collected posts.
We iterate through each post, referred to as the ``main post'' $P_{\text{main}}$. For every $P_{i} \in \mathcal{P}_{\uparrow}$, we attach $P_{i+1}$ as its sole child, producing the path
$P_{0}\!\rightarrow\!\dots\!\rightarrow\!P_{n}\!\rightarrow\!P_{\text{main}}$.
Each $C_{j}\in\mathcal{C}_{\downarrow}$ becomes a child of the main post, giving edges
$P_{\text{main}}\!\rightarrow\!C_{j}$.
Since the same posts can appear in multiple trees, we remove duplicates via URL and recursively union their children.

\paragraph{Extracting Self-contained Samples.}
We then use an LLM to convert trees into samples, as illustrated in~\autoref{fig:1_tree_to_sample}. 
This processing step first identifies the spans of text corresponding to the prompt and discards any unrelated remarks, and performs minor fixups such as combining disjoint spans. 
Next, this step collects any commentary either from the original author or other user replies (e.g., \textit{``amazing result,''} \textit{``didn't work,''} etc.) as a list of community feedback. Finally, the sample is assigned one of three quality labels: ``Benchmark'' (high quality prompts that are coherent and show clear user intent), ``Analysis'' (moderate quality partial prompts or commentary that could not be associated with another sample), or ``Trash'' (off-topic and malformed content that should be discarded).
While we want only the highest quality data for benchmarking, we are also interested in retaining any other relevant data for large-scale analysis. 

\subsection{Multimodal Processing of Posts with Images}\label{subsec:processing_multimodal}
While \autoref{subsec:reconstructing_context} can extract text prompts and community feedback, other metadata requires multimodal processing.
This step marks the input and output images associated with each sample, updates prompts with fill-in-the-blank, and produces new samples by parsing screenshots, addressing three common cases:

\paragraph{Classifying Input vs. Output Images.}
There does not exist a standardized format for marking input and output images. For example, the output image could be the first or the last in a series of images, or there may be irrelevant images that are neither inputs nor outputs.
Nevertheless, users expect viewers to infer this distinction, and thus we use a VLM to make this same inference (see~\autoref{fig:2_classify_images_fib_prompts}).

\paragraph{Completing Fill-in-the-Blank.} A common user behavior is ``fill-in-the-blank'' prompts, where users post a template intended for commenters to infill in the replies. 
Keeping these templates as-is presents a problem, because they are not fully specified and effectively omit the completions that commenters find most interesting. Instead, we use a VLM to reverse-engineer these completions conditioned on the template and the images provided by commenters (see~\autoref{fig:2_classify_images_fib_prompts}).

\paragraph{Extracting Conversation Screenshots.}
Another behavior is sharing screenshots of interactions with ~\oaiimagegen{}, which may contain prompt text, reference images, and image outputs all within the same frame. This is an especially high-quality source of data, since the inputs and outputs are exactly documented without paraphrasing or summary. Extracting the raw data requires a multi-task computer vision system that can localize images to bounding boxes, classify the sub-images as inputs vs. outputs, and detect what is prompt text vs. unrelated conversation. While one could chain together specialized models for each subtask, we instead opt for a more generalizable solution using a VLM.
We first detect these cases with the general multimodal processing prompt, which is then routed to the parsing prompt depicted in~\autoref{fig:3_extract_conversation_screenshot}.
The VLM can not only parse the \oaiimagegen{} interface but also other non-standard layouts, for example side-by-side collages of input and output images. 
For the VLM we opt to use Qwen-2.5-VL~\citep{Qwen2.5-VL}, which is specifically trained for bounding box detection.

\begin{figure}
    \centering
    \includegraphics[width=0.85\linewidth]{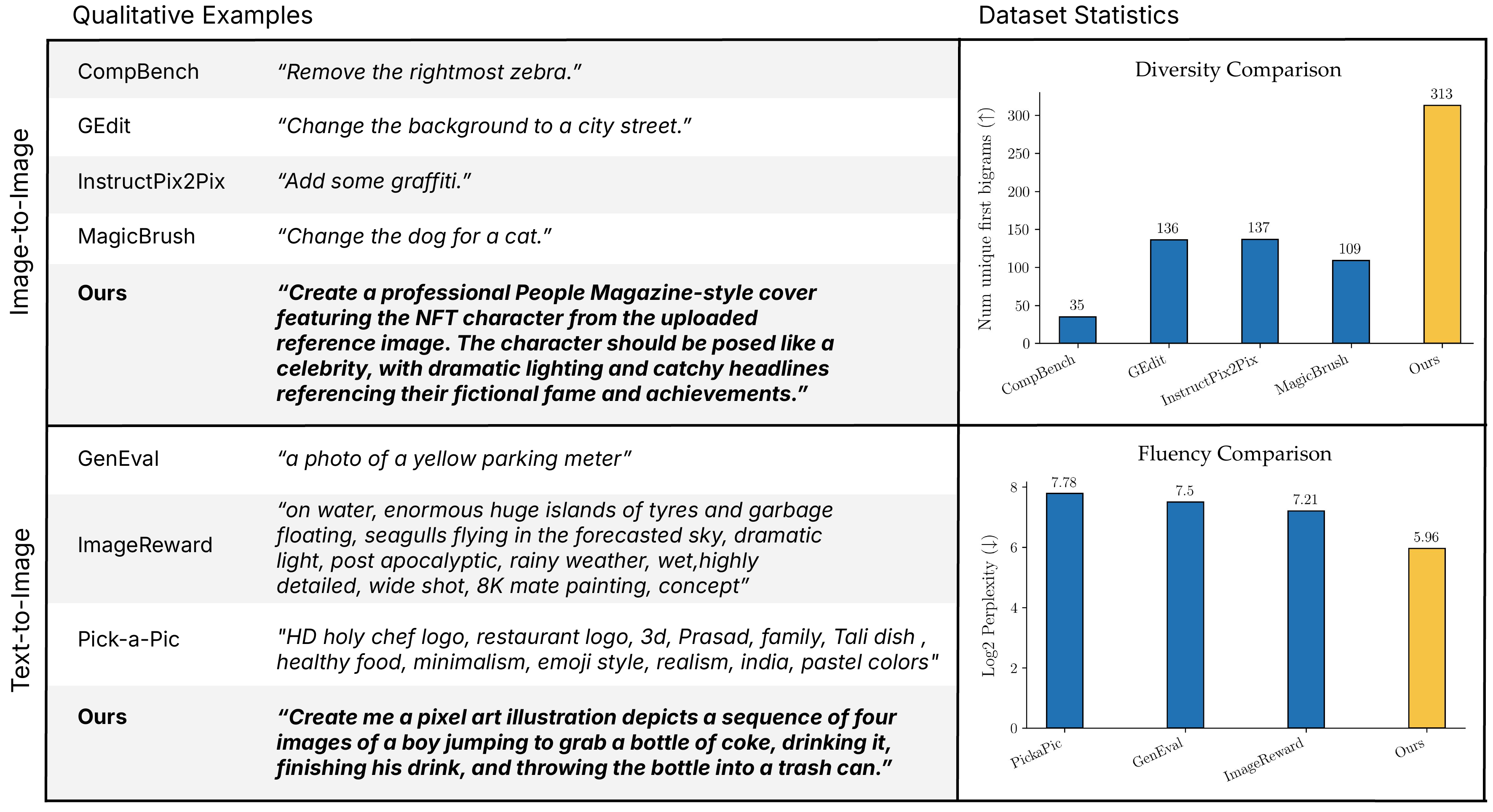}
    \caption{\textbf{Dataset Comparison.} The prompts in \datasetname{} are significantly different from prior benchmarks.
    Top: the image-to-image split is more diverse and complex, with more unique first bigrams. Bottom: the text-to-image split is more fluent, as measured by perplexity under Pythia 12B~\citep{biderman2023pythia}.
    }
    \label{fig:side_by_side}
\end{figure}

\section{\datasetname{}: A Social-Media Post-Release Benchmark}\label{sec:analysis}

We initially run our framework to explore the 4o Image Gen release on Twitter/X.
After the LLM quality filter in \autoref{subsec:reconstructing_context}, we find that 20\% of samples are marked as high-quality (usable for benchmarking) and 66\% are marked as moderate-quality (usable for analysis).
We also apply safety filtering (see more details in \autoref{subsec:ethics}).
This yields 30k total samples for analysis.
For our final benchmark, we limit each split to up to a thousand samples, to keep the downstream costs of benchmarking (generating outputs and rating them) manageable.
We then randomly sample candidates and manually inspect them, flagging low-quality examples for removal or updating their prompts, to ensure that the benchmark is the highest possible quality. This results in an image-to-image split with 710 prompt-image pairs and text-to-image split with 848 prompts.

\paragraph{\datasetname{} Surfaces Diverse and Novel Tasks.} While most benchmarks are limited to templated image editing tasks, such as changing the background, changing the color, adding or replacing an object, as shown in \autoref{fig:side_by_side} (top left), \datasetname{} incorporates several tasks not captured by existing tasks, such as novel view synthesis, image editing that requires cognitive reasoning, virtual try-on, template-based product generation, multi-image subject-driven generation, colorization, image translation, and code-based style transfer (see~\autoref{fig:examples_i2i_qualitative1}-\ref{fig:examples_t2i_qualitative2}). We also can see this diversity effect in the language distribution itself. In \autoref{fig:side_by_side} (top right), we show the unique first-bigrams of each dataset's editing instructions. \datasetname{} also exhibits a substantially larger variety of first bigrams, indicating more diverse instruction types and image operations.

In addition to diversity, \datasetname{} also remains more natural in the language domain. As shown in \autoref{fig:side_by_side} (bottom right), our instructions exhibit consistently lower perplexity, indicating that they align more closely with natural language, and suggesting that users now prefer to interact with generative models using fluent, coherent instructions (compared to previous keyword-centric methods). 

\begin{figure}[t]
    \centering
\includegraphics[width=\linewidth]{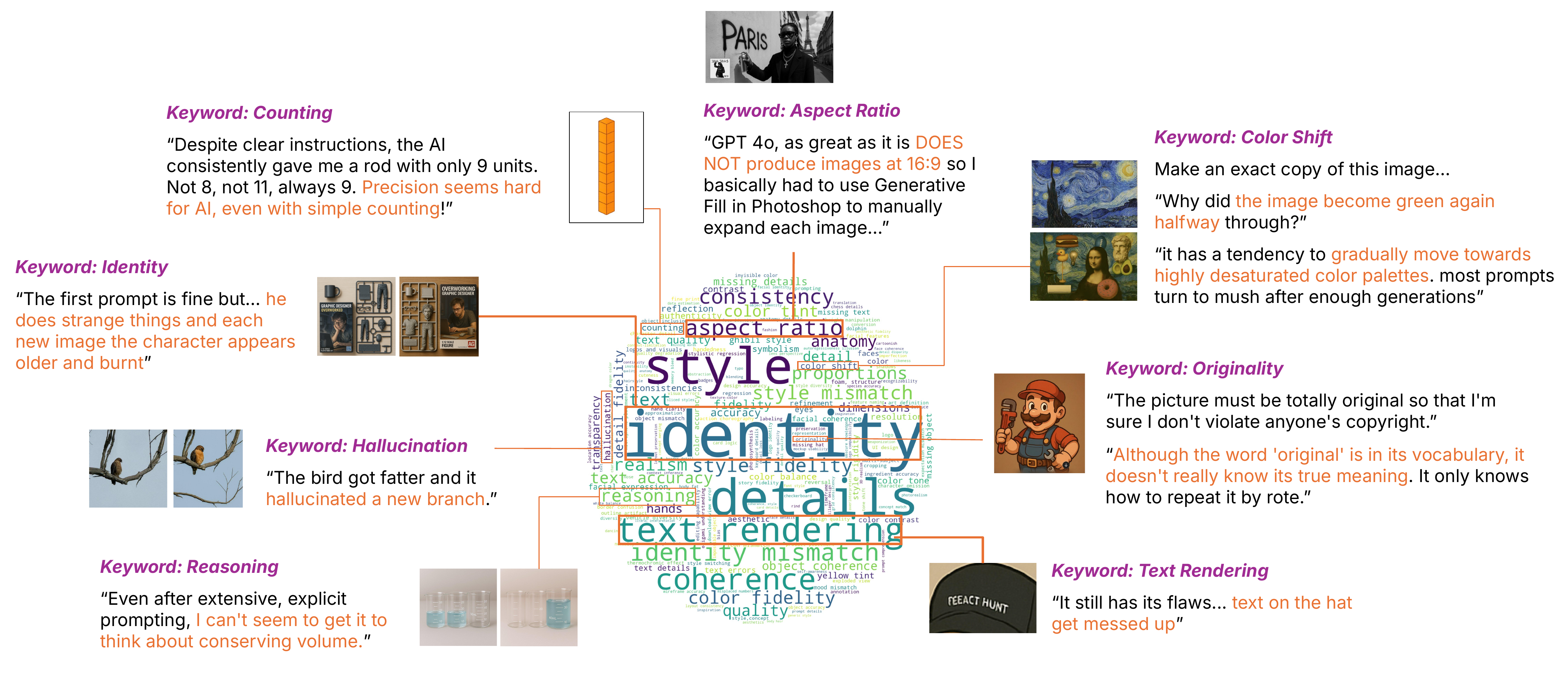}
    \caption{\textbf{Common Failures Observed by Users.} A word cloud of failure cases, derived from community feedback, showing practical capabilities that users care about in real use cases and curiosity-driven tests that reveal deeper model limitations. Common failures include identity shift, color drift, text rendering errors, and style mismatches; more exploratory failures include originality and reasoning about volume.
    }
    \label{fig:qualitative_feedback}
\end{figure}

\paragraph{\datasetname{} Surfaces How Users Interact With Models.} To capture failure modes that users explicitly care about, we first use an LLM to label each piece of community feedback as denoting a success or a failure. Then, for each failure case, the LLM generates a short keyword summary describing the underlying issue (e.g., a failure to render ``a transparent helmet'' correctly will get the keyword ``transparency''). We visualize these keywords as a word cloud and highlight representative cases, as shown in \autoref{fig:qualitative_feedback}.

\begin{figure}[t]
    \centering
    \includegraphics[width=\linewidth]{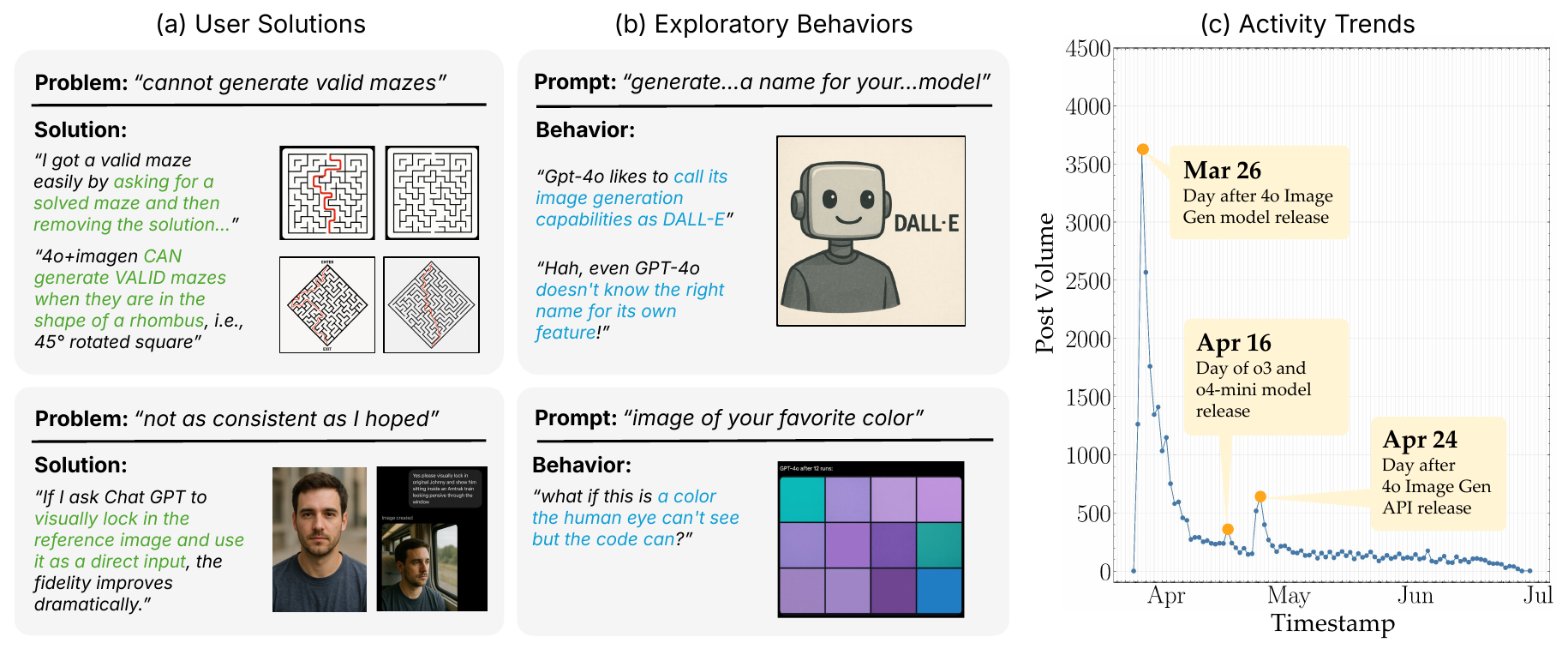}
    \caption{\textbf{How Users Interact with Models.} We depict qualitative examples of (a) user solutions and (b) exploratory behaviors, discovered via community feedback. We also visualize (c) activity trends using the timestamps of collected posts.}
    \label{fig:community_feedback_uses}
\end{figure}

\autoref{fig:qualitative_feedback} shows us that users are generally most sensitive to failure types such as identity shift, color drift, text rendering errors, style mismatches, and aspect ratio inaccuracy. These failure modes reflect practical use cases where users expect reliability and usefulness, and thus indicate areas where improving models would directly enhance satisfaction. Beyond these common issues, \datasetname{} also surfaces corner-case failures that users found interesting. These often come from probing interesting model behaviors, such as reasoning failures in scientific contexts, misunderstandings of concepts such as originality, 
and difficulty with counting.
Such cases reveal deeper limitations of current models and highlight opportunities for future research.
The community feedback from \datasetname{} also reveals practical strategies that users employ to work around model limitations. As shown in \hyperref[fig:community_feedback_uses]{\autoref*{fig:community_feedback_uses}a},  users discuss ways to construct valid mazes or mitigate identity mismatches. In this way, \datasetname{} records crowdsourced prompting solutions to certain issues, which also reflects what users care about, and can help to motivate future model development.

\paragraph{Exploratory Behaviors.}   Interestingly, \datasetname{} also surfaces cases where users explore the model’s behavior itself, rather than pursuing a concrete task.   As shown in \hyperref[fig:community_feedback_uses]{\autoref*{fig:community_feedback_uses}b}, some examples include prompting 4o Image Gen to generate a self-portrait (where it refers to itself as DALL-E) or its favorite color (creating speculation about ``invisible colors'' beyond human vision). These examples illustrate how users collectively probe and reflect on how models behave under novel edge cases, and reveal interesting behaviors that are not captured in standard benchmarks, yet might be of interest to model developers.

\section{\datasetname{} differentiates models}
\label{sec:model_comparison}
\label{sec:setup}

Given our newly curated in-the-wild benchmark, we can now use it to differentiate models.
We evaluate three types of models:
\begin{itemize}[left=1em]
\item \textbf{Unified Models.} To capture the open-source community's progression, we include early models like Anole~\citep{chern2024anole, team2024chameleon} and recent models like Bagel~\citep{deng2025emerging}. We also evaluate proprietary models like 4o Image Gen~\citep{openai2025}, as well as Gemini 2.0 Flash~\citep{comanici2025gemini} and the more recent 2.5 Flash (Nano Banana)~\citep{nanobanana}.
\item \textbf{LLM+Diffusion.} A good baseline for unified models is its most naive implementation: an LLM chained to a diffusion model, where the LLM rewrites the input prompt before diffusion image generation. We follow the best-performing method from~\cite{Zhou_2025_CVPR}, a pipeline with GPT-4o~\citep{oai2024gpt4o} as the LLM and DALL-E 3~\citep{betker2023improving} as the diffusion model.
\item \textbf{Image Editing Models.} Another natural baseline is a specialized image editing model without a sophisticated text encoder. To represent this category, we use Flux Kontext~\citep{batifol2025flux}, which demonstrates state-of-the-art image editing performance.
\end{itemize}

Our overall evaluation metric for the benchmark is head-to-head ``win rate'', a relative rather than absolute metric. Given that our benchmark is composed of in-the-wild prompts that are intrinsically open-ended, it is very challenging to define a notion of ``accuracy.'' The win rate is calculated across all $\binom{n}{2}$ pairwise model comparisons, where each model earns 1 for a win, 0 for a loss, and 0.5 for a tie. Therefore, the final win rate of a model can be interpreted as its average win rate compared with all other models.

\paragraph{Automatic Evaluation.} Due to the cost of collecting human evaluations, we primarily leverage automated evaluation through VLM-as-a-judge. We follow the ``single answer grading'' setup from MT-Bench~\citep{zheng2023judging}. In this setup, a score is directly assigned to each output, then converted into ``pseudo pairwise'' comparisons: for any pair of models, the one with the higher score is treated as the winner. This setup is more scalable as the number of models being evaluated increases, and simplifies the benchmarking process. Furthermore, MT-Bench validates that both true pairwise and pseudo pairwise grading show high agreement with human judgements. To mitigate any biases VLM-as-a-judge might have towards models from the same developer, we ensemble the judgements of three evaluators. We use GPT-4o~\citep{oai2024gpt4o}, Gemini 2.0~\citep{team2023gemini}, and Qwen2.5-VL-32B-Instruct~\citep{Qwen2.5-VL}, then take the majority vote to determine the winner of each model pair.
Following MT-Bench, we instruct the model to produce a chain-of-thought and consider factors like prompt following, fidelity to any reference images, and realism and aesthetics, before producing a score (see~\autoref{fig:0903_autoeval}).

\paragraph{Human Correlation.} As a secondary validation of the automatic evaluator beyond those in \citet{zheng2023judging}, we compare our automated evaluations against against gold label human annotations. Specifically, we present five expert raters with outputs of all 8 models for 200 samples, and ask the annotators to rank the outputs from best to worst for both the text-to-image and image-to-image splits. We found that our VLM-as-a-judge measure correlates weakly, but significantly, with human ratings (GPT: $\tau_b = 0.117_{p=0.0036}$, Gemini: $\tau_b = 0.083_{p=0.0199}$, Qwen: $\tau_b = 0.045_{p=0.1327}$, Human-Human concordance: $W = 0.49_{p < 0.0001}$). While the correlation is positive, and significant for Gemini and GPT, this result suggests that further research into judge models may be necessary for stronger results overall. See \autoref{app:human_evals} for more details.

\subsection{Results}\label{subsec:benchmark_results}

\begin{figure*}[t]
    \centering
    \includegraphics[width=0.9\linewidth]{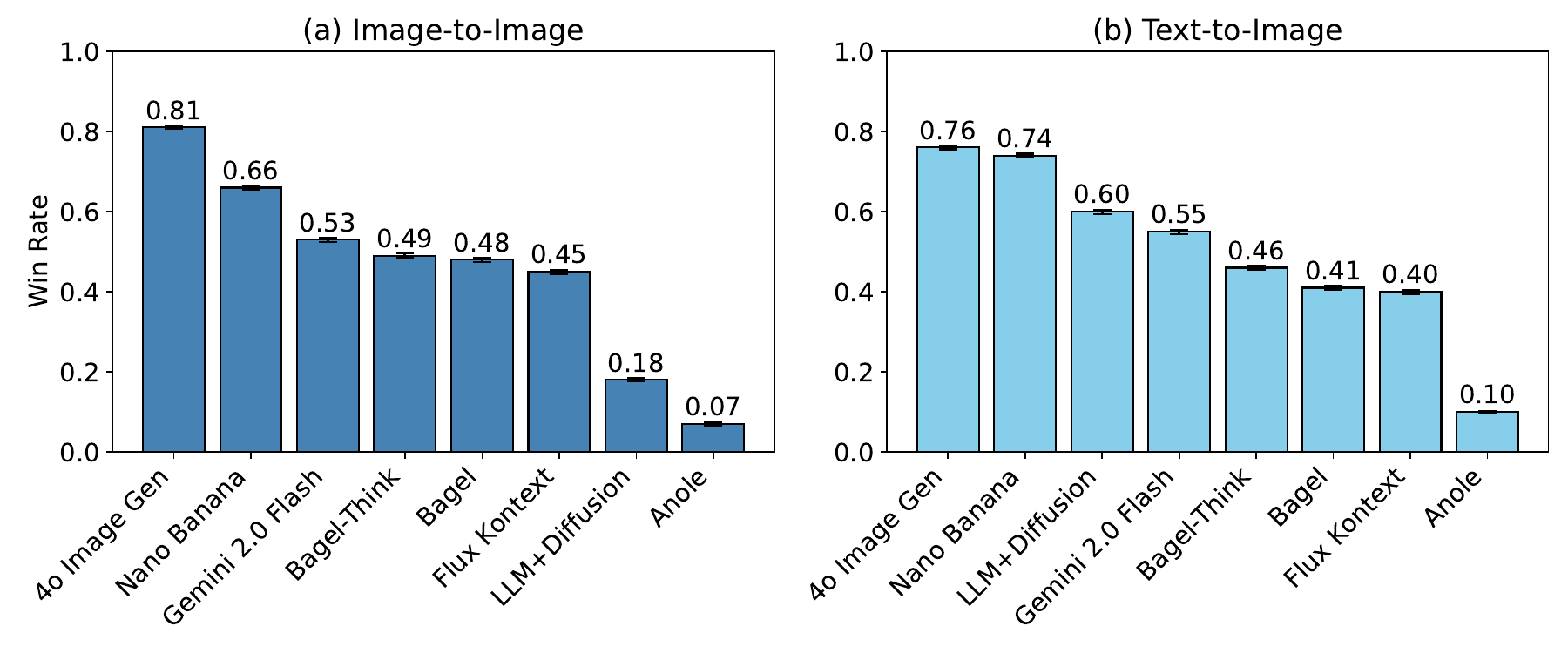}
    \caption{
    \textbf{Overall Evaluation.} We compare a range of unified models, as well as an image editing (Flux Kontext) and shallow fusion (LLM+Diffusion) baseline. We report the win rate, or percentage of pairwise comparisons won. The win rate is calculated automatically with an ensemble of three VLMs-as-a-judge.
    }
    \label{fig:winrate}
\end{figure*}

\begin{figure*}[t]
\includegraphics[width=\linewidth]{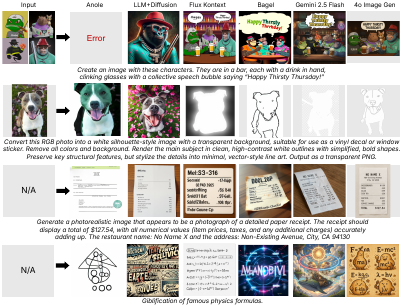}
\caption{\textbf{Qualitative Model Comparison}. Challenging tasks from our benchmark, ranging from translation to multi-concept combination to mathematical reasoning, elicit diverse model responses. We mark samples where the model fails to generate an output as ``Error.''
    }
    \label{fig:qualitative_result}
\end{figure*}

We present the win rate comparison on the image-to-image and text-to-image splits in \autoref{fig:winrate}. Qualitative results of representative models are shown in \autoref{fig:qualitative_result}.
On the image-to-image split (\hyperref[fig:winrate]{\autoref*{fig:winrate}a}), model performance separates into five distinct tiers. First, 4o Image Gen significantly outperforms the other models, followed by Gemini's Nano Banana. Next, Gemini 2.0 Flash, Bagel-Think, Bagel, and Flux Kontext exhibit similar performance. Finally, LLM+Diffusion then Anole perform much worse. We observe similar trends on the text-to-image split (\hyperref[fig:winrate]{\autoref*{fig:winrate}b}), although the gaps are less pronounced and LLM+Diffusion makes a large jump forward in its ranking.

\subsection{Closing the loop with community feedback metrics}\label{subsec:specialized_metrics}

\begin{figure}
    \centering
    \includegraphics[width=\linewidth]{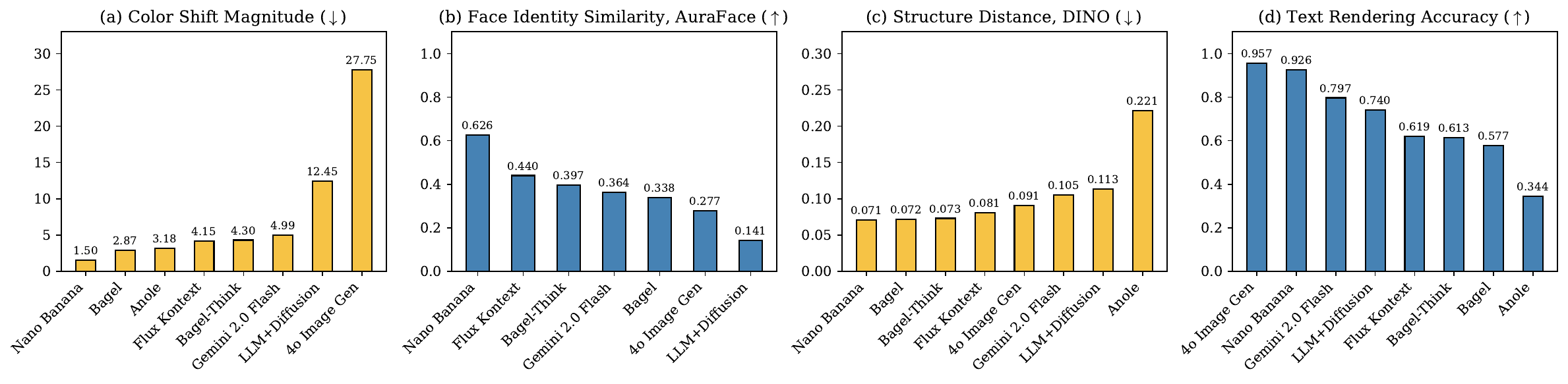}
    \caption{\textbf{Specialized Metrics from Community Feedback.} Based on qualitative community observations, we validate that 4o Image Gen exhibits large shifts in color (a) and face identity (b), moderate shifts in structure distance (c), but superior text rendering accuracy (d).}
    \label{fig:specific_metrics}
\end{figure}

In addition to the automated evaluations in \autoref{sec:model_comparison}, we also wanted to see how community feedback extracted using \datasetname{} could help to differentiate model performance in fine-grained ways. Based on the failure categories extracted by \datasetname{}, and illustrated in~\autoref{fig:qualitative_feedback}, we designed several specialized automated metrics: color shift magnitude, face identity similarity, structure distance, and text rendering accuracy. For each metric, described below, we used an LLM to classify samples where each metric is applicable (\autoref{fig:metrics_classification}), and computed the metric over these samples, with the results presented in \autoref{fig:specific_metrics}.

\textbf{Color Shift Magnitude.}  We quantify the ``yellow tint'' frequently reported in community feedback with a color shift metric, computed as the average difference between the color histogram of the input versus output images.
As shown in \hyperref[fig:specific_metrics]{\autoref*{fig:specific_metrics}a}, 4o Image Gen indeed exhibits the largest color shift. Interestingly, the only other method from the same developer, LLM+Diffusion (implemented with DALLE-3), also exhibits an abnormally large color shift. Users theorize that the yellow tint could be a \textit{``watermarking method, potentially trying to do something kinda fancy with low level pixel encoding.''}

\textbf{Face Identity Similarity.} Community feedback critiques face identity shifts, which we quantify with a face embedding metric. Specifically, we use AuraFace~\citep{deng2019arcface,auraface} to detect faces and extract their embeddings, then select the input-output face pair with the highest cosine similarity. \hyperref[fig:specific_metrics]{\autoref*{fig:specific_metrics}b} confirms user observations that 4o Image Gen struggles with face preservation, which could be attributed to a lossy visual encoder or insufficient identity-oriented training data.

\textbf{Structure Distance.} Users are perceptive towards drift in visual structure, such as object positioning or human pose, which we measure using a DINO-based~\citep{caron2021emerging} structure metric. Following the setup of~\cite{Tumanyan_2023_CVPR}, we compute the Frobenius norm of the Gram matrices derived from DINO key features~\citep{tumanyan2023disentangling} for input-output image pairs.
As expected, methods not specifically trained on image-to-image data (LLM+Diffusion and Anole) perform the worst in structure preservation, as shown in \hyperref[fig:specific_metrics]{\autoref*{fig:specific_metrics}c}. Outside of this category, 4o Image Gen also exhibits non-negligible drift, matching observations that it tends to re-approximate images rather than faithfully copy image structure.

\textbf{Text Rendering Accuracy.} Users are also sensitive 
towards rendered text, which we measure via VLM-as-a-judge.
Unlike OCR-based string matching, VLMs can produce a more holistic score that takes into account factors like legibility in addition to spelling, punctuation, and grammar (see~\autoref{fig:typographic_accuracy}).
\hyperref[fig:specific_metrics]{\autoref*{fig:specific_metrics}d} shows that 4o Image Gen achieves near-perfect text rendering accuracy, consistent with its popularity as a tool for generating infographics and other text-heavy media. 

Together, these results show how community feedback can be systematically translated into targeted quantitative metrics that expose fine-grained tradeoffs across models. Beyond confirming user observations, this approach produces concrete, interpretable signals that can guide model development.

\section{Conclusion}
\label{sec:conclusion}
In this work, we introduced \datasetname{}, the first framework for evaluating image generation in alignment with emerging, real-world use cases of modern image models. Applied to social media posts about GPT-4o Image Gen, \datasetname{} surfaces novel use cases not captured by prior benchmarks, differentiates proprietary from open-source models, and motivates targeted metrics grounded in common failure cases such as text rendering. As both models and user needs evolve, so too must the benchmarks that guide their development.

\newpage
\subsection*{Ethics Statement}\label{subsec:ethics}
In this work, we primarily study discussion of \oaiimagegen{} on \twitter{}, a public social media platform where users voluntarily share content, for academic research purposes. Our collection process implicitly benefits from existing moderation systems: \twitter{} removes posts that violate its content policies~\citep{x_post_policy}, and ChatGPT refuses to generate images that violate its usage guidelines~\citep{openai_image_policy_2025}.
For this reason, the collected posts are relatively benign, as illustrated by qualitative examples from our dataset (see~\autoref{app:additional_examples}).
We also take additional steps to remove potentially harmful material.
For all samples, we applied LLama-Guard-4-12B~\citep{llamaguard4}, a multimodal safety classifier designed to safeguard according to the MLCommons hazards taxonomy~\citep{ghosh2025ailuminateintroducingv10ai}. We then removed any samples with text or images flagged to contain any of its hazard categories, such as violent, sexual, hateful, or privacy-violating content.
To minimize privacy risk, we also manually exclude input images that plausibly depict anyone under eighteen.

\subsection*{Acknowledgements}\label{subsec:ack}
We thank Stephanie Fu, Michelle Li, and Alexander Pan for their helpful feedback. We also thank the folks at Stochastic Labs for previewing early prototypes of this work. Finally, we extend a special thank you to Lisa Dunlap for entertaining many extensive discussions on evaluations.

\bibliography{iclr2026_conference}

\begin{thebibliography}{42}
\providecommand{\natexlab}[1]{#1}
\providecommand{\url}[1]{\texttt{#1}}
\expandafter\ifx\csname urlstyle\endcsname\relax
  \providecommand{\doi}[1]{doi: #1}\else
  \providecommand{\doi}{doi: \begingroup \urlstyle{rm}\Url}\fi

\bibitem[Andrieu et~al.(2023)Andrieu, Cohen-Boulakia, Couceiro, Denise, and Pierrot]{andrieu4353494unifying}
Pierre Andrieu, Sarah Cohen-Boulakia, Miguel Couceiro, Alain Denise, and Adeline Pierrot.
\newblock A unifying rank aggregation framework to suitably and efficiently aggregate any kind of rankings.
\newblock \emph{International Journal of Approximate Reasoning}, 162:\penalty0 109035, 2023.
\newblock ISSN 0888-613X.
\newblock \doi{https://doi.org/10.1016/j.ijar.2023.109035}.
\newblock URL \url{https://www.sciencedirect.com/science/article/pii/S0888613X23001664}.

\bibitem[Bai et~al.(2025)Bai, Chen, Liu, Wang, Ge, Song, Dang, Wang, Wang, Tang, Zhong, Zhu, Yang, Li, Wan, Wang, Ding, Fu, Xu, Ye, Zhang, Xie, Cheng, Zhang, Yang, Xu, and Lin]{Qwen2.5-VL}
Shuai Bai, Keqin Chen, Xuejing Liu, Jialin Wang, Wenbin Ge, Sibo Song, Kai Dang, Peng Wang, Shijie Wang, Jun Tang, Humen Zhong, Yuanzhi Zhu, Mingkun Yang, Zhaohai Li, Jianqiang Wan, Pengfei Wang, Wei Ding, Zheren Fu, Yiheng Xu, Jiabo Ye, Xi~Zhang, Tianbao Xie, Zesen Cheng, Hang Zhang, Zhibo Yang, Haiyang Xu, and Junyang Lin.
\newblock Qwen2.5-vl technical report.
\newblock \emph{arXiv preprint arXiv:2502.13923}, 2025.

\bibitem[Batifol et~al.(2025)Batifol, Blattmann, Boesel, Consul, Diagne, Dockhorn, English, English, Esser, Kulal, et~al.]{batifol2025flux}
Stephen Batifol, Andreas Blattmann, Frederic Boesel, Saksham Consul, Cyril Diagne, Tim Dockhorn, Jack English, Zion English, Patrick Esser, Sumith Kulal, et~al.
\newblock Flux. 1 kontext: Flow matching for in-context image generation and editing in latent space.
\newblock \emph{arXiv e-prints}, pp.\  arXiv--2506, 2025.

\bibitem[Betker et~al.(2023)Betker, Goh, Jing, Brooks, Wang, Li, Ouyang, Zhuang, Lee, Guo, et~al.]{betker2023improving}
James Betker, Gabriel Goh, Li~Jing, Tim Brooks, Jianfeng Wang, Linjie Li, Long Ouyang, Juntang Zhuang, Joyce Lee, Yufei Guo, et~al.
\newblock Improving image generation with better captions (2023).
\newblock \emph{URL https://cdn. openai. com/papers/dall-e-3. pdf}, 6, 2023.

\bibitem[Biderman et~al.(2023)Biderman, Schoelkopf, Anthony, Bradley, O’Brien, Hallahan, Khan, Purohit, Prashanth, Raff, et~al.]{biderman2023pythia}
Stella Biderman, Hailey Schoelkopf, Quentin~Gregory Anthony, Herbie Bradley, Kyle O’Brien, Eric Hallahan, Mohammad~Aflah Khan, Shivanshu Purohit, USVSN~Sai Prashanth, Edward Raff, et~al.
\newblock Pythia: A suite for analyzing large language models across training and scaling.
\newblock In \emph{International Conference on Machine Learning}, pp.\  2397--2430. PMLR, 2023.

\bibitem[Brooks et~al.(2023)Brooks, Holynski, and Efros]{brooks2023instructpix2pix}
Tim Brooks, Aleksander Holynski, and Alexei~A Efros.
\newblock Instructpix2pix: Learning to follow image editing instructions.
\newblock In \emph{Proceedings of the IEEE/CVF conference on computer vision and pattern recognition}, pp.\  18392--18402, 2023.

\bibitem[Caron et~al.(2021)Caron, Touvron, Misra, J\'egou, Mairal, Bojanowski, and Joulin]{caron2021emerging}
Mathilde Caron, Hugo Touvron, Ishan Misra, Herv\'e J\'egou, Julien Mairal, Piotr Bojanowski, and Armand Joulin.
\newblock Emerging properties in self-supervised vision transformers.
\newblock In \emph{Proceedings of the International Conference on Computer Vision (ICCV)}, 2021.

\bibitem[Chern et~al.(2024)Chern, Su, Ma, and Liu]{chern2024anole}
Ethan Chern, Jiadi Su, Yan Ma, and Pengfei Liu.
\newblock Anole: An open, autoregressive, native large multimodal models for interleaved image-text generation.
\newblock \emph{arXiv preprint arXiv:2407.06135}, 2024.

\bibitem[Chiang et~al.(2024)Chiang, Zheng, Sheng, Angelopoulos, Li, Li, Zhu, Zhang, Jordan, Gonzalez, et~al.]{chiang2024chatbot}
Wei-Lin Chiang, Lianmin Zheng, Ying Sheng, Anastasios~Nikolas Angelopoulos, Tianle Li, Dacheng Li, Banghua Zhu, Hao Zhang, Michael Jordan, Joseph~E Gonzalez, et~al.
\newblock Chatbot arena: An open platform for evaluating llms by human preference.
\newblock In \emph{Forty-first International Conference on Machine Learning}, 2024.

\bibitem[Comanici et~al.(2025)Comanici, Bieber, Schaekermann, Pasupat, Sachdeva, Dhillon, Blistein, Ram, Zhang, Rosen, et~al.]{comanici2025gemini}
Gheorghe Comanici, Eric Bieber, Mike Schaekermann, Ice Pasupat, Noveen Sachdeva, Inderjit Dhillon, Marcel Blistein, Ori Ram, Dan Zhang, Evan Rosen, et~al.
\newblock Gemini 2.5: Pushing the frontier with advanced reasoning, multimodality, long context, and next generation agentic capabilities.
\newblock \emph{arXiv preprint arXiv:2507.06261}, 2025.

\bibitem[{ComfyUI Wiki}(2025)]{comfyui_prompt_basic_2025}
{ComfyUI Wiki}.
\newblock Basic syntax tips for comfyui prompt writing, 2025.
\newblock URL \url{https://comfyui-wiki.com/en/tutorial/basic/stable-diffusion-prompt-basic}.

\bibitem[Deng et~al.(2025)Deng, Zhu, Li, Gou, Li, Wang, Zhong, Yu, Nie, Song, et~al.]{deng2025emerging}
Chaorui Deng, Deyao Zhu, Kunchang Li, Chenhui Gou, Feng Li, Zeyu Wang, Shu Zhong, Weihao Yu, Xiaonan Nie, Ziang Song, et~al.
\newblock Emerging properties in unified multimodal pretraining.
\newblock \emph{arXiv preprint arXiv:2505.14683}, 2025.

\bibitem[Deng et~al.(2019)Deng, Guo, Xue, and Zafeiriou]{deng2019arcface}
Jiankang Deng, Jia Guo, Niannan Xue, and Stefanos Zafeiriou.
\newblock Arcface: Additive angular margin loss for deep face recognition.
\newblock In \emph{Proceedings of the IEEE/CVF conference on computer vision and pattern recognition}, pp.\  4690--4699, 2019.

\bibitem[{fal}(2025)]{auraface}
{fal}.
\newblock Auraface, 2025.
\newblock URL \url{https://huggingface.co/fal/AuraFace-v1}.

\bibitem[{Gemini}(2025)]{nanobanana}
{Gemini}.
\newblock Nano banana, 2025.
\newblock URL \url{https://gemini.google/overview/image-generation}.

\bibitem[Ghosh et~al.(2023)Ghosh, Hajishirzi, and Schmidt]{ghosh2023geneval}
Dhruba Ghosh, Hannaneh Hajishirzi, and Ludwig Schmidt.
\newblock Geneval: An object-focused framework for evaluating text-to-image alignment.
\newblock \emph{Advances in Neural Information Processing Systems}, 36:\penalty0 52132--52152, 2023.

\bibitem[Ghosh et~al.(2025)Ghosh, Frase, Williams, Luger, Röttger, Barez, McGregor, Fricklas, Kumar, Feuillade-Montixi, Bollacker, Friedrich, Tsang, Vidgen, Parrish, Knotz, Presani, Bennion, Boston, Kuniavsky, Hutiri, Ezick, Salem, Sahay, Goswami, Gohar, Huang, Sarin, Alhajjar, Chen, Eng, Manjusha, Mehta, Long, Emani, Vidra, Rukundo, Shahbazi, Chen, Ghosh, Thangarasa, Peigné, Singh, Bartolo, Krishna, Akhtar, Gold, Coleman, Oala, Tashev, Imperial, Russ, Kunapuli, Miailhe, Delaunay, Radharapu, Shinde, Tuesday, Dutta, Grabb, Gangavarapu, Sahay, Gangavarapu, Schramowski, Singam, David, Han, Mammen, Prabhakar, Kovatchev, Weiss, Ahmed, Manyeki, Madireddy, Khomh, Zhdanov, Baumann, Vasan, Yang, Mougn, Varghese, Chinoy, Jitendar, Maskey, Hardgrove, Li, Gupta, Joswin, Mai, Kumar, Patlak, Lu, Alessi, Balija, Gu, Sullivan, Gealy, Lavrisa, Goel, Mattson, Liang, and Vanschoren]{ghosh2025ailuminateintroducingv10ai}
Shaona Ghosh, Heather Frase, Adina Williams, Sarah Luger, Paul Röttger, Fazl Barez, Sean McGregor, Kenneth Fricklas, Mala Kumar, Quentin Feuillade-Montixi, Kurt Bollacker, Felix Friedrich, Ryan Tsang, Bertie Vidgen, Alicia Parrish, Chris Knotz, Eleonora Presani, Jonathan Bennion, Marisa~Ferrara Boston, Mike Kuniavsky, Wiebke Hutiri, James Ezick, Malek~Ben Salem, Rajat Sahay, Sujata Goswami, Usman Gohar, Ben Huang, Supheakmungkol Sarin, Elie Alhajjar, Canyu Chen, Roman Eng, Kashyap~Ramanandula Manjusha, Virendra Mehta, Eileen Long, Murali Emani, Natan Vidra, Benjamin Rukundo, Abolfazl Shahbazi, Kongtao Chen, Rajat Ghosh, Vithursan Thangarasa, Pierre Peigné, Abhinav Singh, Max Bartolo, Satyapriya Krishna, Mubashara Akhtar, Rafael Gold, Cody Coleman, Luis Oala, Vassil Tashev, Joseph~Marvin Imperial, Amy Russ, Sasidhar Kunapuli, Nicolas Miailhe, Julien Delaunay, Bhaktipriya Radharapu, Rajat Shinde, Tuesday, Debojyoti Dutta, Declan Grabb, Ananya Gangavarapu, Saurav Sahay, Agasthya Gangavarapu, Patrick
  Schramowski, Stephen Singam, Tom David, Xudong Han, Priyanka~Mary Mammen, Tarunima Prabhakar, Venelin Kovatchev, Rebecca Weiss, Ahmed Ahmed, Kelvin~N. Manyeki, Sandeep Madireddy, Foutse Khomh, Fedor Zhdanov, Joachim Baumann, Nina Vasan, Xianjun Yang, Carlos Mougn, Jibin~Rajan Varghese, Hussain Chinoy, Seshakrishna Jitendar, Manil Maskey, Claire~V. Hardgrove, Tianhao Li, Aakash Gupta, Emil Joswin, Yifan Mai, Shachi~H Kumar, Cigdem Patlak, Kevin Lu, Vincent Alessi, Sree~Bhargavi Balija, Chenhe Gu, Robert Sullivan, James Gealy, Matt Lavrisa, James Goel, Peter Mattson, Percy Liang, and Joaquin Vanschoren.
\newblock Ailuminate: Introducing v1.0 of the ai risk and reliability benchmark from mlcommons, 2025.
\newblock URL \url{https://arxiv.org/abs/2503.05731}.

\bibitem[Huang et~al.(2023)Huang, Sun, Xie, Li, and Liu]{huang2023t2i}
Kaiyi Huang, Kaiyue Sun, Enze Xie, Zhenguo Li, and Xihui Liu.
\newblock T2i-compbench: A comprehensive benchmark for open-world compositional text-to-image generation.
\newblock \emph{Advances in Neural Information Processing Systems}, 36:\penalty0 78723--78747, 2023.

\bibitem[Hui et~al.(2024)Hui, Yang, Zhao, Shi, Wang, Wang, Zhou, and Xie]{hui2024hq}
Mude Hui, Siwei Yang, Bingchen Zhao, Yichun Shi, Heng Wang, Peng Wang, Yuyin Zhou, and Cihang Xie.
\newblock Hq-edit: A high-quality dataset for instruction-based image editing.
\newblock \emph{arXiv preprint arXiv:2404.09990}, 2024.

\bibitem[Jiang et~al.(2025)Jiang, Jiang, Yang, Liu, Tsang, and Shou]{jiang2025balanced}
Yuxin Jiang, Liming Jiang, Shuai Yang, Jia-Wei Liu, Ivor Tsang, and Mike~Zheng Shou.
\newblock Balanced image stylization with style matching score.
\newblock \emph{arXiv preprint arXiv:2503.07601}, 2025.

\bibitem[Kirstain et~al.(2023)Kirstain, Polyak, Singer, Matiana, Penna, and Levy]{kirstain2023pick}
Yuval Kirstain, Adam Polyak, Uriel Singer, Shahbuland Matiana, Joe Penna, and Omer Levy.
\newblock Pick-a-pic: An open dataset of user preferences for text-to-image generation.
\newblock \emph{Advances in neural information processing systems}, 36:\penalty0 36652--36663, 2023.

\bibitem[Lee et~al.(2023)Lee, Yasunaga, Meng, Mai, Park, Gupta, Zhang, Narayanan, Teufel, Bellagente, et~al.]{lee2023holistic}
Tony Lee, Michihiro Yasunaga, Chenlin Meng, Yifan Mai, Joon~Sung Park, Agrim Gupta, Yunzhi Zhang, Deepak Narayanan, Hannah Teufel, Marco Bellagente, et~al.
\newblock Holistic evaluation of text-to-image models.
\newblock \emph{Advances in Neural Information Processing Systems}, 36:\penalty0 69981--70011, 2023.

\bibitem[Liu et~al.(2025)Liu, Han, Xing, Yin, Wang, Cheng, Liao, Wang, Fu, Han, et~al.]{liu2025step1x}
Shiyu Liu, Yucheng Han, Peng Xing, Fukun Yin, Rui Wang, Wei Cheng, Jiaqi Liao, Yingming Wang, Honghao Fu, Chunrui Han, et~al.
\newblock Step1x-edit: A practical framework for general image editing.
\newblock \emph{arXiv preprint arXiv:2504.17761}, 2025.

\bibitem[{Llama Team}()]{llamaguard4}
{Llama Team}.
\newblock Llama guard 4 model card.
\newblock URL \url{https://huggingface.co/meta-llama/Llama-Guard-4-12B}.

\bibitem[OpenAI(2024)]{oai2024gpt4o}
OpenAI.
\newblock Gpt-4o system card, 2024.
\newblock URL \url{https://openai.com/index/gpt-4o-system-card}.

\bibitem[{OpenAI}(2025{\natexlab{a}})]{openai2025}
{OpenAI}.
\newblock Addendum to gpt-4o system card: Native image generation, 2025{\natexlab{a}}.
\newblock Accessed: 2025-08-24.

\bibitem[{OpenAI}(2025{\natexlab{b}})]{openai_image_policy_2025}
{OpenAI}.
\newblock Creating images and videos in line with our policies, 2025{\natexlab{b}}.
\newblock URL \url{https://openai.com/policies/creating-images-and-videos-in-line-with-our-policies/}.

\bibitem[Radford et~al.(2021)Radford, Kim, Hallacy, Ramesh, Goh, Agarwal, Sastry, Askell, Mishkin, Clark, et~al.]{radford2021learning}
Alec Radford, Jong~Wook Kim, Chris Hallacy, Aditya Ramesh, Gabriel Goh, Sandhini Agarwal, Girish Sastry, Amanda Askell, Pamela Mishkin, Jack Clark, et~al.
\newblock Learning transferable visual models from natural language supervision.
\newblock In \emph{International conference on machine learning}, pp.\  8748--8763. PmLR, 2021.

\bibitem[Rombach et~al.(2022)Rombach, Blattmann, Lorenz, Esser, and Ommer]{rombach2022high}
Robin Rombach, Andreas Blattmann, Dominik Lorenz, Patrick Esser, and Bj{\"o}rn Ommer.
\newblock High-resolution image synthesis with latent diffusion models.
\newblock In \emph{Proceedings of the IEEE/CVF conference on computer vision and pattern recognition}, pp.\  10684--10695, 2022.

\bibitem[Sheynin et~al.(2024)Sheynin, Polyak, Singer, Kirstain, Zohar, Ashual, Parikh, and Taigman]{sheynin2024emu}
Shelly Sheynin, Adam Polyak, Uriel Singer, Yuval Kirstain, Amit Zohar, Oron Ashual, Devi Parikh, and Yaniv Taigman.
\newblock Emu edit: Precise image editing via recognition and generation tasks.
\newblock In \emph{Proceedings of the IEEE/CVF Conference on Computer Vision and Pattern Recognition}, pp.\  8871--8879, 2024.

\bibitem[Team(2024)]{team2024chameleon}
Chameleon Team.
\newblock Chameleon: Mixed-modal early-fusion foundation models.
\newblock \emph{arXiv preprint arXiv:2405.09818}, 2024.

\bibitem[Team et~al.(2023)Team, Anil, Borgeaud, Alayrac, Yu, Soricut, Schalkwyk, Dai, Hauth, Millican, et~al.]{team2023gemini}
Gemini Team, Rohan Anil, Sebastian Borgeaud, Jean-Baptiste Alayrac, Jiahui Yu, Radu Soricut, Johan Schalkwyk, Andrew~M Dai, Anja Hauth, Katie Millican, et~al.
\newblock Gemini: a family of highly capable multimodal models.
\newblock \emph{arXiv preprint arXiv:2312.11805}, 2023.

\bibitem[Tumanyan et~al.(2023{\natexlab{a}})Tumanyan, Bar-Tal, Amir, Bagon, and Dekel]{tumanyan2023disentangling}
Narek Tumanyan, Omer Bar-Tal, Shir Amir, Shai Bagon, and Tali Dekel.
\newblock Disentangling structure and appearance in vit feature space.
\newblock \emph{ACM Trans. Graph.}, nov 2023{\natexlab{a}}.
\newblock ISSN 0730-0301.
\newblock \doi{10.1145/3630096}.
\newblock URL \url{https://doi.org/10.1145/3630096}.

\bibitem[Tumanyan et~al.(2023{\natexlab{b}})Tumanyan, Geyer, Bagon, and Dekel]{Tumanyan_2023_CVPR}
Narek Tumanyan, Michal Geyer, Shai Bagon, and Tali Dekel.
\newblock Plug-and-play diffusion features for text-driven image-to-image translation.
\newblock In \emph{Proceedings of the IEEE/CVF Conference on Computer Vision and Pattern Recognition (CVPR)}, pp.\  1921--1930, June 2023{\natexlab{b}}.

\bibitem[Wang et~al.(2023)Wang, Saharia, Montgomery, Pont-Tuset, Noy, Pellegrini, Onoe, Laszlo, Fleet, Soricut, et~al.]{wang2023imagen}
Su~Wang, Chitwan Saharia, Ceslee Montgomery, Jordi Pont-Tuset, Shai Noy, Stefano Pellegrini, Yasumasa Onoe, Sarah Laszlo, David~J Fleet, Radu Soricut, et~al.
\newblock Imagen editor and editbench: Advancing and evaluating text-guided image inpainting.
\newblock In \emph{Proceedings of the IEEE/CVF conference on computer vision and pattern recognition}, pp.\  18359--18369, 2023.

\bibitem[Wang et~al.(2022)Wang, Montoya, Munechika, Yang, Hoover, and Chau]{wang2022diffusiondb}
Zijie~J Wang, Evan Montoya, David Munechika, Haoyang Yang, Benjamin Hoover, and Duen~Horng Chau.
\newblock Diffusiondb: A large-scale prompt gallery dataset for text-to-image generative models.
\newblock \emph{arXiv preprint arXiv:2210.14896}, 2022.

\bibitem[{X Help Center}(2025)]{x_post_policy}
{X Help Center}.
\newblock The x rules, 2025.
\newblock URL \url{https://help.x.com/en/rules-and-policies/x-rules}.

\bibitem[Xu et~al.(2023)Xu, Liu, Wu, Tong, Li, Ding, Tang, and Dong]{xu2023imagereward}
Jiazheng Xu, Xiao Liu, Yuchen Wu, Yuxuan Tong, Qinkai Li, Ming Ding, Jie Tang, and Yuxiao Dong.
\newblock Imagereward: Learning and evaluating human preferences for text-to-image generation.
\newblock \emph{Advances in Neural Information Processing Systems}, 36:\penalty0 15903--15935, 2023.

\bibitem[Zhang et~al.(2023{\natexlab{a}})Zhang, Mo, Chen, Sun, and Su]{Zhang2023MagicBrush}
Kai Zhang, Lingbo Mo, Wenhu Chen, Huan Sun, and Yu~Su.
\newblock Magicbrush: A manually annotated dataset for instruction-guided image editing.
\newblock In \emph{Advances in Neural Information Processing Systems}, 2023{\natexlab{a}}.

\bibitem[Zhang et~al.(2023{\natexlab{b}})Zhang, Yang, Feng, Qin, Chen, Yu, Chen, Wang, Savarese, Ermon, Xiong, and Xu]{zhang2023hive}
Shu Zhang, Xinyi Yang, Yihao Feng, Can Qin, Chia-Chih Chen, Ning Yu, Zeyuan Chen, Huan Wang, Silvio Savarese, Stefano Ermon, Caiming Xiong, and Ran Xu.
\newblock Hive: Harnessing human feedback for instructional visual editing.
\newblock \emph{arXiv preprint arXiv:2303.09618}, 2023{\natexlab{b}}.

\bibitem[Zheng et~al.(2023)Zheng, Chiang, Sheng, Zhuang, Wu, Zhuang, Lin, Li, Li, Xing, Zhang, Gonzalez, and Stoica]{zheng2023judging}
Lianmin Zheng, Wei-Lin Chiang, Ying Sheng, Siyuan Zhuang, Zhanghao Wu, Yonghao Zhuang, Zi~Lin, Zhuohan Li, Dacheng Li, Eric Xing, Hao Zhang, Joseph~E. Gonzalez, and Ion Stoica.
\newblock Judging {LLM}-as-a-judge with {MT}-bench and chatbot arena.
\newblock In \emph{Thirty-seventh Conference on Neural Information Processing Systems Datasets and Benchmarks Track}, 2023.
\newblock URL \url{https://openreview.net/forum?id=uccHPGDlao}.

\bibitem[Zhou et~al.(2025)Zhou, Peng, Song, Li, Xu, Yang, Guo, Zhang, Lin, He, Zhao, Liu, Li, Xie, Chang, Qiao, Shao, and Zhang]{Zhou_2025_CVPR}
Pengfei Zhou, Xiaopeng Peng, Jiajun Song, Chuanhao Li, Zhaopan Xu, Yue Yang, Ziyao Guo, Hao Zhang, Yuqi Lin, Yefei He, Lirui Zhao, Shuo Liu, Tianhua Li, Yuxuan Xie, Xiaojun Chang, Yu~Qiao, Wenqi Shao, and Kaipeng Zhang.
\newblock Opening: A comprehensive benchmark for judging open-ended interleaved image-text generation.
\newblock In \emph{Proceedings of the IEEE/CVF Conference on Computer Vision and Pattern Recognition (CVPR)}, pp.\  56--66, June 2025.

\end{thebibliography}
\bibliographystyle{iclr2026_conference}

\newpage
\appendix
\counterwithin{figure}{section}
\counterwithin{table}{section}

\section*{Appendix}
The Appendix is organized as follows:
\begin{itemize}[left=1em]
\item \autoref{app:limitations} discusses the limitations of our framework.
\item \autoref{app:additional_examples} gives some additional qualitative examples from \datasetname{}.
\item \autoref{app:data_collection} provides the prompts for, and some additional information on, the data collection pipeline.
\item \autoref{app:eval_metrics} provides the prompts used for automatic evaluation with VLM-as-a-judge.
\item \autoref{app:human_evals} discusses the results and process used for our human validation of VLM-as-a-judge.
\item \autoref{app:llm_disclosure} discusses the use of LLMs in the preparation of this manuscript.
\end{itemize}

\section{Limitations}
\label{app:limitations}
While \datasetname{} is diverse and markedly distinct from prior benchmarks, it may not be representative of \textit{all} possible user queries. First, there is a bias towards certain topics; for example there is an unusually large number of requests for \textit{``Ghibli style''} due to social media trends. Second, users are more likely to post examples where \oaiimagegen{} succeeds rather than fails, which affects the distribution of tested capabilities. However, these quirks are inherent to crowdsourced datasets; DiffusionDB~\citep{wang2022diffusiondb} is similarly biased towards \textit{``artstation style''} and keyword lists favorable towards Stable Diffusion. As such, these benchmarks should be viewed as comparisons to the current best model in the community consciousness, rather than arbiters of the ``universally best'' model for any user query. For this reason, we present not only a benchmark but also a reproducible framework, which can be re-run as soon as a new model with new capabilities is released, or as soon as community interests change.

\clearpage
\section{Additional Examples}
\label{app:additional_examples}

\begin{figure}
    \centering
    \includegraphics[width=\linewidth]{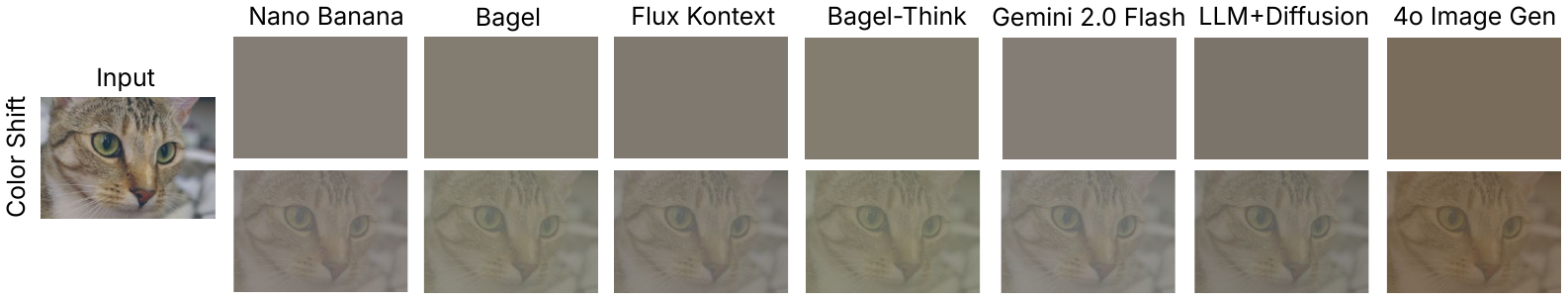}
    \caption{\textbf{Per-Model Average Color Histogram.} For each model, we compute the average color histogram of its outputs on the image-to-image split (top), then overlay it on top of a real image as a visual aid (bottom). Evidently, 4o Image Gen exhibits a substantial yellow tint.}
    \label{fig:color_shift}
\end{figure}

\begin{figure}
    \centering
    \includegraphics[width=\linewidth]{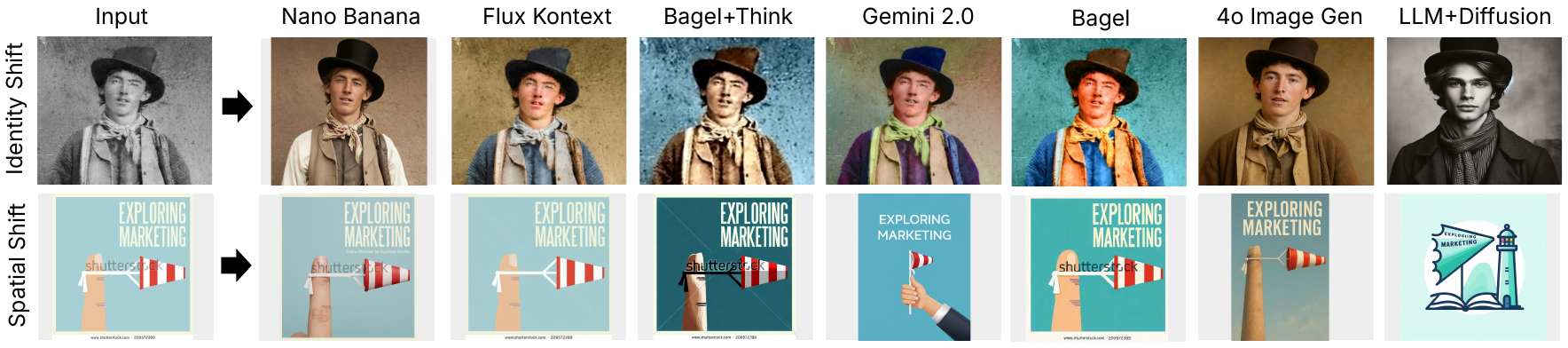}
    \caption{\textbf{Qualitative Comparison of Identity and Spatial Shift.} Given the prompt \textit{``Billy the Kid cleaned up and colorized from the famous photo of him''} each model retains the input identity to varying degrees. For the prompt \textit{``giving it a fresh twist with a more detailed, realistic touch''} each model retains the input image's spatial layout by a different amount.
    }
    \label{fig:identity_spatial_shift}
\end{figure}

\paragraph{Examples Illustrating Specialized Metrics.} In~\autoref{fig:color_shift} and~\autoref{fig:identity_spatial_shift} we display examples illustrating the range of drift in color, identity, and spatial structure across different methods.

\begin{figure}
    \centering
    \includegraphics[width=0.95\linewidth]{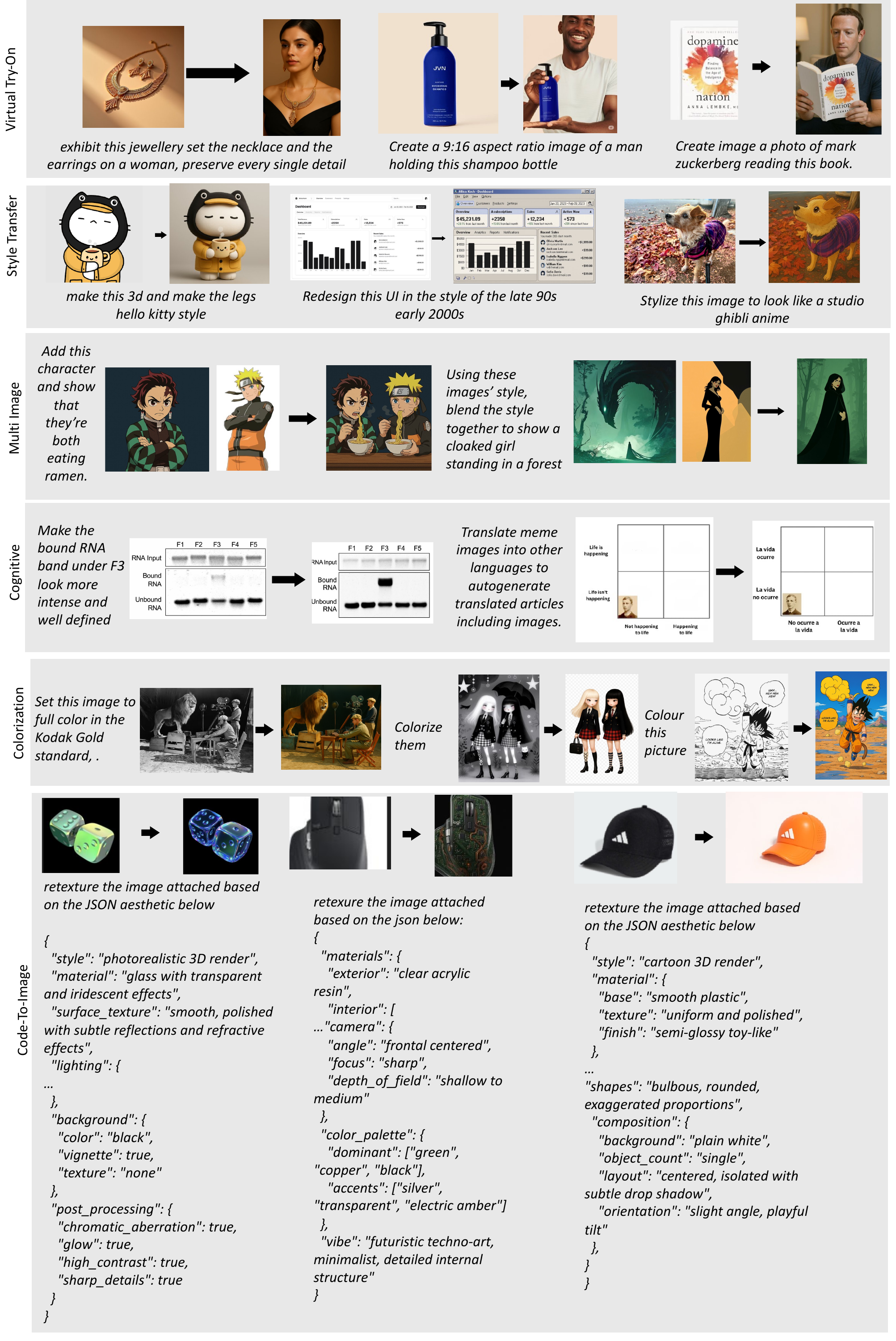}
    \caption{Image-to-image examples from \datasetname{}.}
    \label{fig:examples_i2i_qualitative1}
\end{figure}

\begin{figure}
    \centering
    \includegraphics[width=0.95\linewidth]{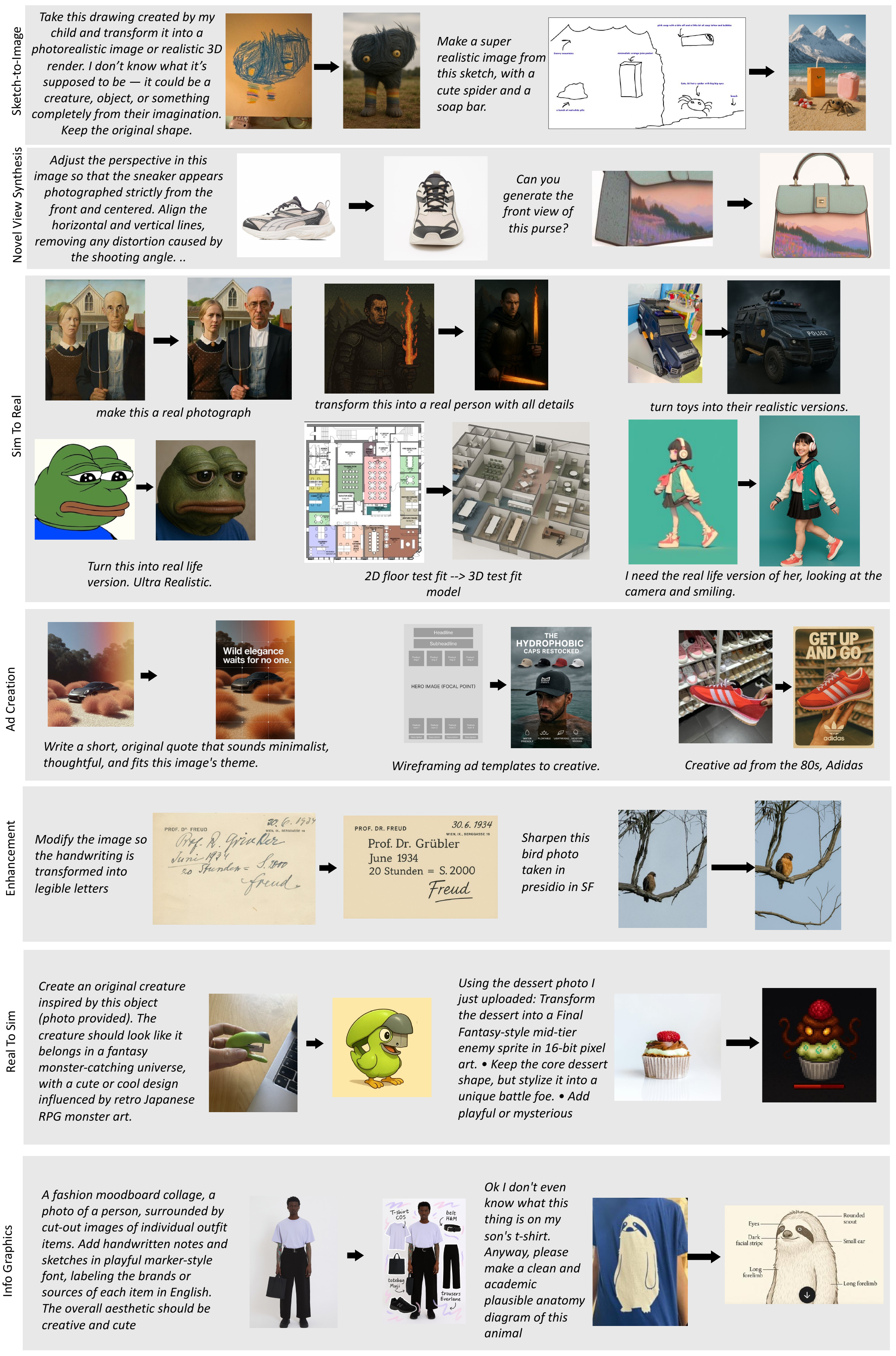}
    \caption{Image-to-image examples from \datasetname{}, continued.}
    \label{fig:examples_i2i_qualitative2}
\end{figure}

\begin{figure}
    \centering
    \includegraphics[width=0.8\linewidth]{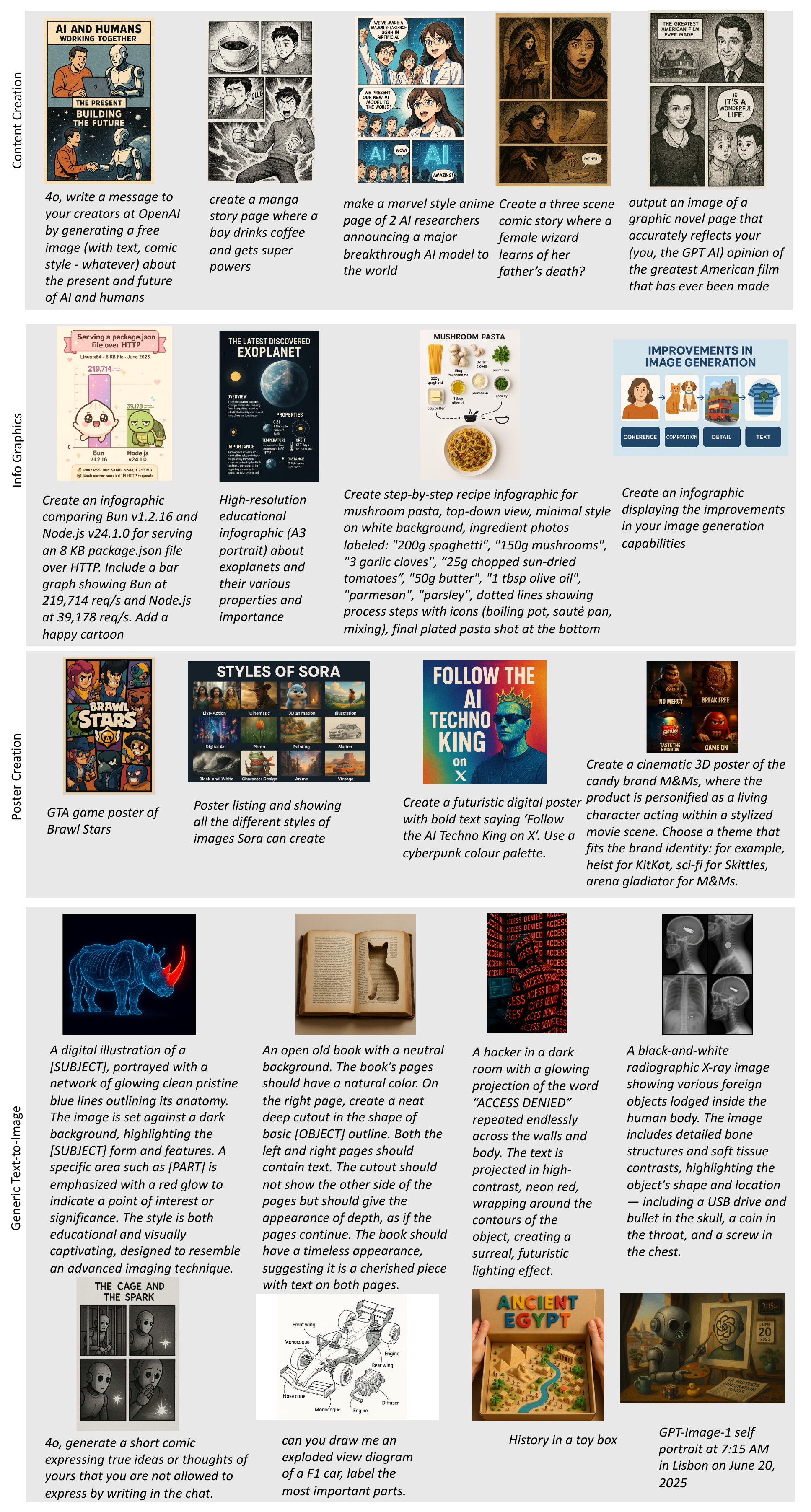}
    \caption{Text-to-image examples from \datasetname{}.}
    \label{fig:examples_t2i_qualitative1}
\end{figure}

\begin{figure}
    \centering
    \includegraphics[width=0.8\linewidth]{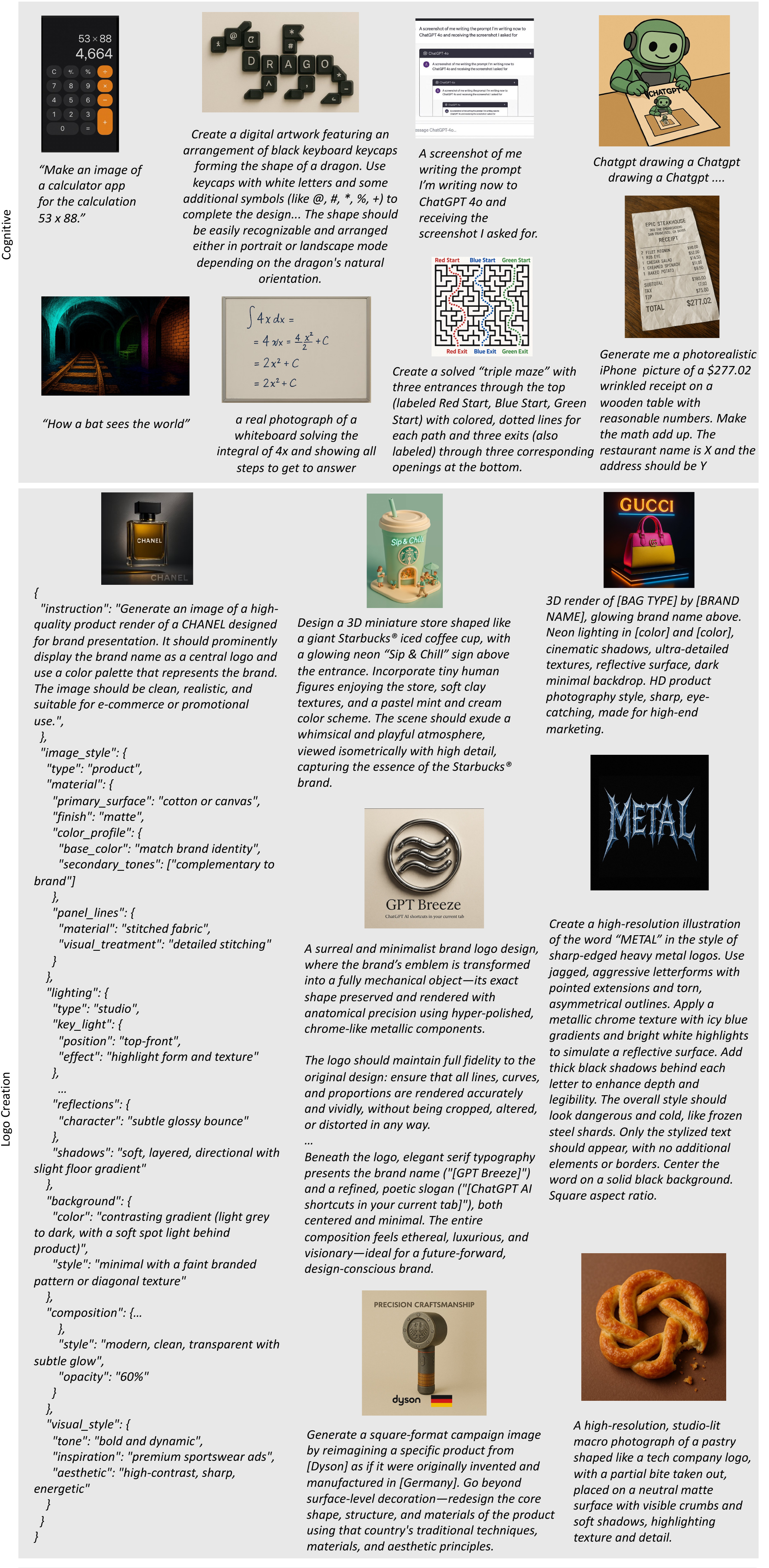}
    \caption{Text-to-image examples from \datasetname{}.}
    \label{fig:examples_t2i_qualitative2}
\end{figure}

\paragraph{Qualitative Examples from~\datasetname{}.} In \autoref{fig:examples_i2i_qualitative1}, ~\autoref{fig:examples_i2i_qualitative2},~\autoref{fig:examples_t2i_qualitative1}, ~\autoref{fig:examples_t2i_qualitative2} we highlight further qualitative examples surfaced through the \datasetname{} framework. These examples demonstrate the breadth of tasks that naturally arise from community use of current image generation models, going beyond traditional templated image editing tasks and crowdsourced prompts centered around Stable Diffusion.

\clearpage

\clearpage

\clearpage

\clearpage
\section{Human Ranking \& Correlation with LLMs}
\label{app:human_evals}

\begin{figure}
    \centering
    \includegraphics[width=\linewidth]{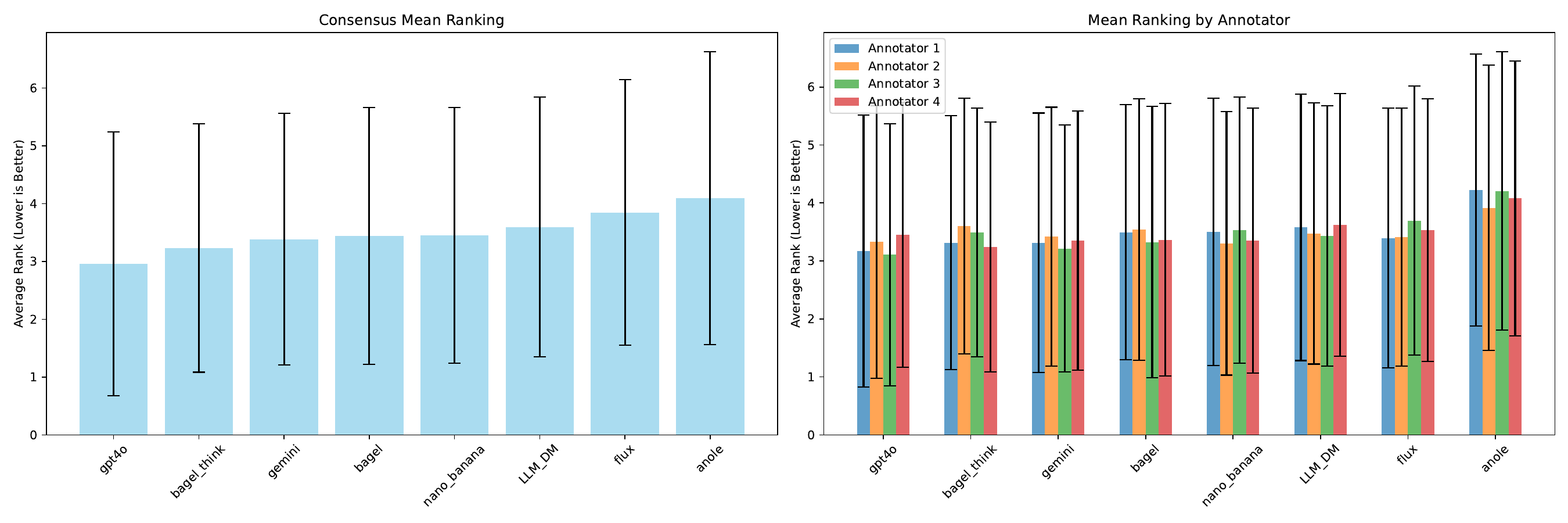}
    \caption{The consensus ranking, and mean ranking by annotator for each of the models. As we can see, because of the limited size of the annotation sets, the standard deviations of the bars is quite high, meaning that we can draw very few conclusions about model performance overall from the human data.}
    \label{fig:human_rank}
\end{figure}

To evaluate the performance of our LLM as a judge models, we performed a limited human evaluation using five expert raters in our group. Each rater fully ranked each of the 8 models over 200 samples (100 from the text-only split, and 100 from the interleaved split), flagging any samples that were impossible to rank fairly. \autoref{fig:human_rank} shows the aggregate of the rankings for each model. 

While the number of annotations is somewhat low for determining model performance, we wanted to understand if the samples that we collected (200) could show significant results in terms of model ordering. To do so, we first ran a Friedman Test on the rankings, and found that with $p < 0.001$ there was a significant difference between the means of the rankings. To determine which pairs are actually significant, we further performed a Dunn's test for significant pairwise differences, and found that after Bonferroni correction, only 8/28 model pairs were significant, shown in \autoref{tab:signficant_human_diff}.

\begin{table}[t]
    \centering
    \scriptsize
    \caption{Significant model differences from human evaluation. We can see that even from our relatively limited human evaluation, anole and LLM\_DM under-perform most models, primarily due to the image-editing split, where both perform quite poorly.}
    \begin{tabularx}{\linewidth}{lXcccc}
    \toprule
    \textbf{Model A} & \textbf{Model B} & \textbf{Z-Statistic} & \textbf{P-Value (raw)} & \textbf{P-Value (Bonf.)} & \textbf{Signficance} \\
    \midrule
        anole   & gpt4o       & 7.429 & 0.000000 & 0.000000 & *** \\
        anole   & nano\_banana & 6.218 & 0.000000 & 0.000000 & *** \\
        anole   & flux        & 6.175 & 0.000000 & 0.000000 & *** \\
        anole   & bagel\_think & 6.127 & 0.000000 & 0.000000 & *** \\
        anole   & gemini      & 5.755 & 0.000000 & 0.000000 & *** \\
        anole   & bagel       & 4.377 & 0.000012 & 0.000337 & *** \\
        LLM\_DM & anole       & 4.011 & 0.000061 & 0.001694 & **  \\
        LLM\_DM & gpt4o       & 3.418 & 0.000631 & 0.017675 & *   \\
        bagel   & gpt4o       & 3.052 & 0.002274 & 0.063676 & -  \\
        LLM\_DM & nano\_banana & 2.207 & 0.027314 & 0.764788 & -  \\
        LLM\_DM & flux        & 2.164 & 0.030461 & 0.852901 & -  \\
        LLM\_DM & bagel\_think & 2.117 & 0.034287 & 0.960036 & -  \\
    \bottomrule
    \end{tabularx}
    \label{tab:signficant_human_diff}
\end{table}

To compute annotator-LLM agreement, we first constructed a consensus ranking for the human raters using the Kemeny-Young method \citep{andrieu4353494unifying}. The split rankings were then merged, giving a total of 200 samples. The LLM as a judge methods produce a single floating point score for each sample. In order to compare the methods, we construct a ranking for each LLM judge from these scores, breaking ties randomly. We then computed Kendall's $\tau_b$ with each of the LLM judges, giving us the presented results in \autoref{sec:setup}, GPT: $\tau_b = 0.117_{p=0.0036}$, Gemini: $\tau_b = 0.083_{p=0.0199}$, Qwen: $\tau_b = 0.045_{p=0.1327}$. We notice here that while GPT and Gemini both have weak, but significant correlations, Qwen does not correlate significantly with human judgment across the raters, and is thus, unlikely to serve as a strong judge for human performance. Interestingly, however, the LLMs correlate with each other. We computed the pearson-r correlation between the scores of pairs of annotators: GPT $\leftrightarrow$ Gemini: $r=0.575_{p=0}$, Gemini $\leftrightarrow$ Qwen: $r=0.627_{p=0}$,  GPT $\leftrightarrow$ Qwen: $r=0.480_{p=0}$. 

\begin{figure}
    \centering
    \includegraphics[width=\linewidth]{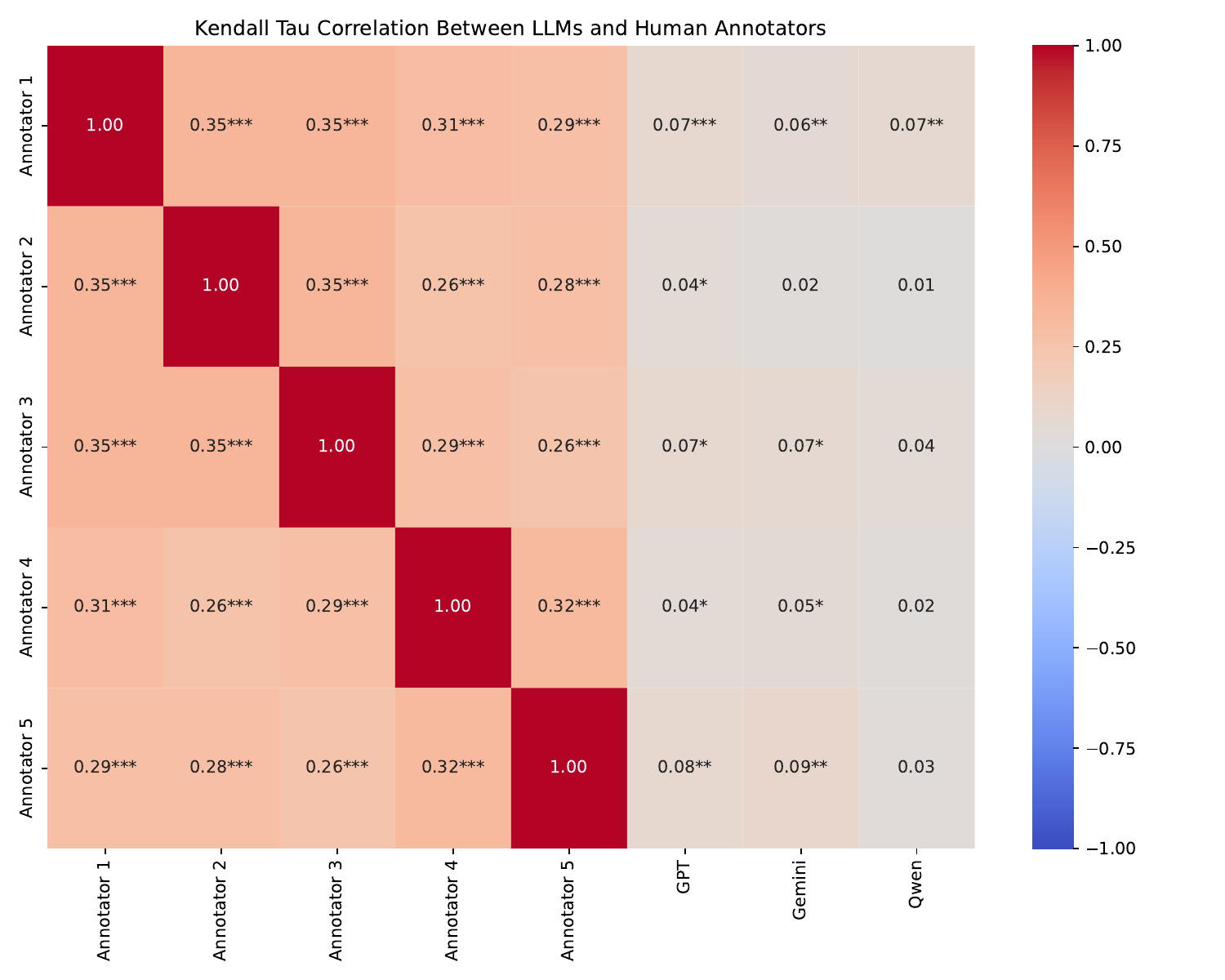}
    \caption{Kendall's $\tau_b$ for pairs of individual human raters and LLM judges (without consensus ranking). We can see that human annotators have fairly high inter-rater correlation, while LLM judges have slight positive correlations, with most correlations signficant among them. In the figure, $*** \to p < 0.001$,  $** \to p < 0.01$, $* \to p < 0.05$.}
    \label{fig:human_tau}
\end{figure}

Another interesting finding is that the Kendall's tau-b for each of our raters differed dramatically. \autoref{fig:human_tau} shows the correlation between each of the human raters, and each of the LLM judges independently (no consensus ranking). We can see that while three of our raters (graduate students on the project) have high inter-annotator correlation, two other raters (undergraduates on the project) have notably different preferences, some of which correlate better with models than others. 

\clearpage
\section{Data Collection Pipeline}
\label{app:data_collection}

\paragraph{Dataset Fields.} In~\autoref{tab:data_fields_full} we display the full set of metadata associated with each sample after running the entire~\datasetname{} framework. 

\begin{figure*}
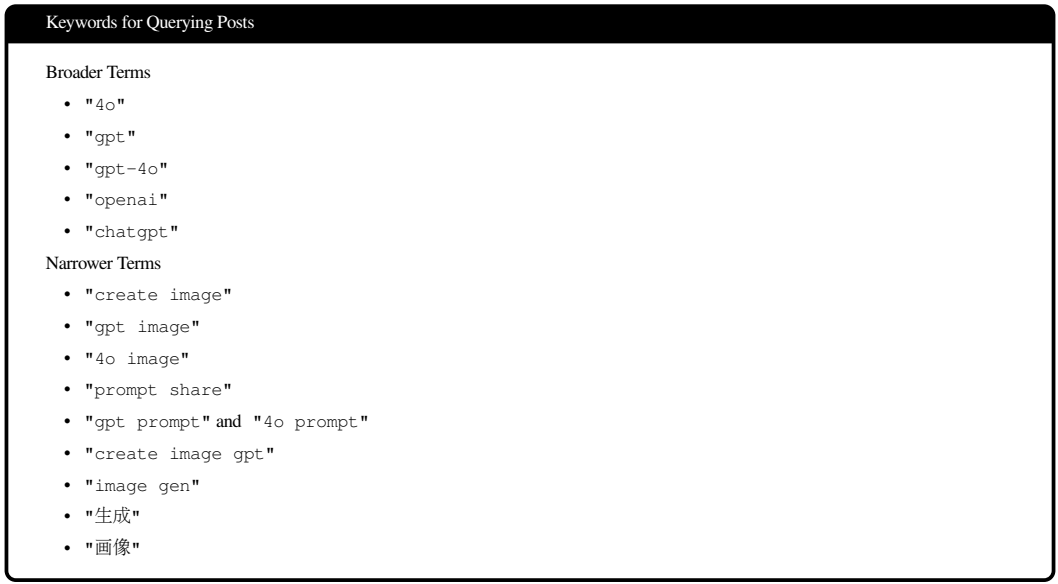

\centering
\scriptsize
\begin{tcolorbox}[title=Keywords for Querying Posts, colframe=black!100!white, colback=white]

Broader Terms
\begin{itemize}[left=1em]
    \item \texttt{"4o"}
    \item \texttt{"gpt"}
    \item \texttt{"gpt-4o"} 
    \item \texttt{"openai"}
    \item \texttt{"chatgpt"}
\end{itemize}

Narrower Terms
\begin{itemize}[left=1em]
\item \texttt{"create image"} 
    \item \texttt{"gpt image"} 
    \item \texttt{"4o image"}
    \item \texttt{"prompt share"}
    \item \texttt{"gpt prompt"} and \texttt{"4o prompt"} 
    \item \texttt{"create image gpt"} 
    \item \texttt{"image gen"} 
    \item \texttt{"\begin{CJK*}{UTF8}{gbsn}生成\end{CJK*}"}
    \item \texttt{"\begin{CJK*}{UTF8}{gbsn}画像\end{CJK*}"}
\end{itemize}

\end{tcolorbox}
\caption{Keywords used to query posts, described in~\autoref{subsec:maximizing_volume}. For the initial two weeks following the 4o Image Gen release we favor volume: we query more generic terms, over daily intervals. In later weeks we favor precision: we query more targeted terms often used for sharing image generation results, over weekly intervals. To increase coverage in foreign languages, we also use calligraphic keywords applicable to Chinese and Japanese, while the alphanumeric keywords are sufficient to also cover Romance languages like Spanish and French.
}
\label{fig:query_keywords}
\end{figure*}

\begin{figure*}
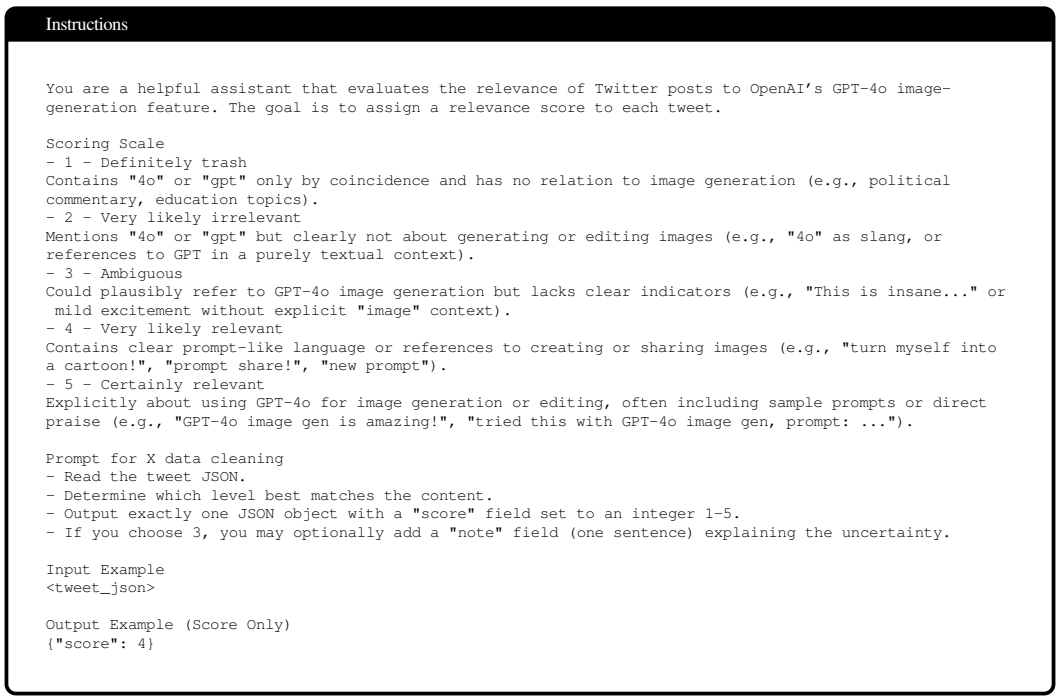

\centering
\scriptsize
\begin{tcolorbox}[colback=white, colframe=black!75!black, title=Instructions]
\begin{lstlisting}
You are a helpful assistant that evaluates the relevance of Twitter posts to OpenAI's GPT-4o image-generation feature. The goal is to assign a relevance score to each tweet.

Scoring Scale
- 1 - Definitely trash
Contains "4o" or "gpt" only by coincidence and has no relation to image generation (e.g., political commentary, education topics).
- 2 - Very likely irrelevant
Mentions "4o" or "gpt" but clearly not about generating or editing images (e.g., "4o" as slang, or references to GPT in a purely textual context).
- 3 - Ambiguous
Could plausibly refer to GPT-4o image generation but lacks clear indicators (e.g., "This is insane..." or mild excitement without explicit "image" context).
- 4 - Very likely relevant
Contains clear prompt-like language or references to creating or sharing images (e.g., "turn myself into a cartoon!", "prompt share!", "new prompt").
- 5 - Certainly relevant
Explicitly about using GPT-4o for image generation or editing, often including sample prompts or direct praise (e.g., "GPT-4o image gen is amazing!", "tried this with GPT-4o image gen, prompt: ...").

Prompt for X data cleaning
- Read the tweet JSON.
- Determine which level best matches the content.
- Output exactly one JSON object with a "score" field set to an integer 1-5.
- If you choose 3, you may optionally add a "note" field (one sentence) explaining the uncertainty.

Input Example
<tweet_json>

Output Example (Score Only)
{"score": 4}
\end{lstlisting}
\end{tcolorbox}
\caption{Prompt for relevance filtering after raw data collection with GPT-4o~\citep{oai2024gpt4o}, described in~\autoref{subsec:maximizing_volume}.}
\label{fig:0_coarse_filter}
\end{figure*}

\paragraph{Data Collection and Processing.} 
We discuss the design of keywords for querying posts in~\autoref{fig:query_keywords}. We also display the prompts used for each step of data processing in the~\datasetname{} framework, including relevance filtering (\autoref{fig:0_coarse_filter}), extracting trees into samples (\autoref{fig:1_tree_to_sample}), multimodal processing (\autoref{fig:2_classify_images_fib_prompts}), and parsing screenshots of conversations (\autoref{fig:3_extract_conversation_screenshot}).

\begin{table}
\scriptsize
\centering
\caption{
Full sample metadata before and after processing with~\datasetname{}.
}
\label{tab:data_fields_full}
\begin{tabular}{p{0.24\linewidth} p{0.70\linewidth}}
\toprule
\multicolumn{2}{l}{\textbf{Raw Retrieved Fields}} \\
\midrule
\texttt{text} & Post text content. \\
\texttt{timestamp} & Posting time of the tweet. \\
\texttt{replies\_above} & Context tweets obtained by scrolling upward in the thread. \\
\texttt{keyword} & Search keyword used to retrieve the post. \\
\texttt{url} & Direct URL of the post. \\
\texttt{author} & Username of the post author. \\
\texttt{image\_urls} & List of image URLs with associated ALT text. \\
\texttt{replies\_below} & Replies obtained by scrolling downward in the thread. \\
\texttt{engagement} & Engagement statistics of the post, including likes/views/reposts/bookmarks. \\

\midrule
\multicolumn{2}{l}{\textbf{Post-Processed Fields}} \\
\midrule
\texttt{id} & Unique identifier for the sample. \\
\texttt{post\_id} & The identifier of the primary post. \\
\texttt{prompt} & User instruction or query text extracted from the original post. \\
\texttt{prompt\_modified} & Boolean flag indicating whether the prompt was manually edited during cleaning. \\
\texttt{quality} & Label describing the intended use of the sample (e.g., ``Benchmark''). \\
\texttt{community\_feedback} & List of replies or comments from other users, each stored with its post\_id and feedback text. \\
\texttt{input\_images} & References to images that are additional input(s) with the prompt.\\
\texttt{output\_images} & References to images that are the output(s) of the prompt.\\
\texttt{is\_screenshot} & Whether the sample's input and/or output images need to be extracted from a screenshot. \\
\texttt{input\_bboxs} & Bounding boxes to crop the screenshot to obtain the true input images.\\
\texttt{output\_bboxs} & Bounding boxes to crop the screenshot to obtain the true output images.\\
\bottomrule
\end{tabular}
\end{table}

\clearpage

\clearpage
\begin{figure*}
\centering
\scriptsize
\begin{tcolorbox}[colback=white, colframe=black!75!black, title=Instructions]
\begin{lstlisting}
You are an extractor of multimodal prompts for image generation.

You will be given a JSON that represents a Twitter post and its reply tree. Each post in the tree may contain an image generation prompt; your job is to extract them into unique samples.
For every input, try to extract at least one sample rather than returning an empty list. We want to extract as many samples as possible, and use a quality score for filtering.
Please output a JSON list of samples in the format ```json [...]```.

## Post to Prompt
Each sample should include the following keys:
{"prompt": <str>, "prompt_modified": <bool>, "post_urls": <list of strs>}

To extract the "prompt":
- Identify each post that discusses a unique image generation task. Set "prompt" as the post text that describes this task. Be very broad in the definition of "prompt"; any instruction, description, comment, or question that hints at an image generation task is fine.
- Make a new sample for every new prompt, even if it is a slight modification of another sample's prompt.
- Try to extract the prompt from the post text exactly, without modification. You may modify the prompt when the modification is obvious, for example, piecing together text from multiple posts or filling in placeholder text. Set the flag "prompt_modified" to True or False accordingly.
- Omissions of text should not be considered as modifications; you should omit statements that are obviously not part of the prompt.
- Many main posts say something like "Prompt Below" or "Prompt in Next Comment"; this means that the tree is likely to have a really good sample, and the prompt needs to be found in the replies.

To determine "post_urls":
- For each "prompt", set "post_urls" to the urls of posts in the tree that likely contain images that are related inputs or outputs for that prompt, which you can determine from the post text.
- Order "post_urls" by importance; the first url should contain the main task information.
- Many replies use a similar prompt as the root post and attach an output image. These should be grouped in the "post_urls" of the main post. Try to infer if this is happening from the reply text.
- If the reply's text indicates a new task, it should be a new sample. If the reply's text indicates it is irrelevant to image generation, it should be omitted. If the reply contains no text and an image, it should be included in "post_urls" so that it can be further processed later.
- Each url/post should appear at most once; images should not be duplicated across samples.

## Quality Score
Each sample should also classify the prompt quality:
{"quality": <str>}

To classify "quality":
- Classify the quality as one of the following categories: ["Benchmark", "Analysis", "Trash"].
- "Benchmark" are the highest quality prompts, which instruct a single coherent image generation task, that can be used for benchmarking. Be fairly strict about the quality.
- "Analysis" are moderate quality prompts that are not in "prompt" format, which are often comments or questions relevant to image generators but do not query a specific task, and are still usable for analysis.
- "Trash" are low quality prompts that have no clear task or are clearly irrelevant. Our focus is on OpenAI's gpt-image-1 or 4o image generation; if the post clearly uses another model or platform like DALL-E, Stable Diffusion, some video generator, etc. it should be classified as "Trash".
- Make sure to collect as many "Analysis" samples as possible, while maintaining relevancy. For these samples, set "prompt" to be the relevant text or commentary about image generation.

## Community Feedback
Each sample should contain a list of community feedback:
{"community_feedback": [{"post_url": <str>, "feedback": <str>}, ...]}

To extract "community_feedback":
- For each post in the tree, determine whether it discusses the sample's success / quality (e.g., "really cool", "does not work", etc.).
- If a post obviously does not have feedback, do not include it.
- The feedback may come from the main post's author or from other authors in the replies.
- Include the full feedback text without modification such that there is sufficient context, but also omit obviously irrelevant text.
- Each url/post should appear at most once in the "community_feedback"; feedback should not be duplicated across samples.

json_post_tree: <tree>
extracted:
\end{lstlisting}
\end{tcolorbox}
\caption{Prompt for tree-to-sample extraction with GPT-4o~\citep{oai2024gpt4o}, described in~\autoref{subsec:reconstructing_context}.}
\label{fig:1_tree_to_sample}
\end{figure*}
\begin{figure*}
\centering
\scriptsize
\begin{tcolorbox}[colback=white, colframe=black!75!black, title=Instructions]
\begin{lstlisting}

You are an extractor of multimodal prompts for image generation.

Your job is to process input-output image pairs from raw user prompts for image generation collected from Twitter. 
You will be given a prompt, a dictionary mapping image ids to images, and a dictionary mapping image ids to post urls.
Please output a JSON list of samples in the format ```json [...]```.

## Image Classification
Each sample should include the following keys, which categorize images as inputs or outputs:
{"inputs": <list of ids>, "outputs": <list of ids>, "post_urls": <list of strs>}

To classify "inputs" and "outputs":
- Inputs, combined with the prompt, should produce a fully specified and coherent image generation task.
- Outputs should be plausible results given the inputs and prompt.
- You may encounter tasks like text-to-image generation (no inputs), image editing (one input), or multi-image conditioned generation (multiple inputs).
- Set "post_urls" the list of urls associated with the inputs and outputs. Order "post_urls" by importance; the first url should contain the main task information.

General rules:
- Each category is mutually exclusive. Each image should not be assigned to multiple categories.
- Some images are low quality and irrelevant to any task. They should not be assigned to any category.
- Some samples are low quality where it is not possible to extract any coherent task. Simply return an empty dictionary {}.
- If there are no relevant images, assign an empty list [] to the category.
- If an image is duplicated, use the smaller index as the id and ignore the others.
- Each id should appear at most once. Each post_url should appear at most once.

## Fill in the Blank
The input prompt may be a "fill in the blank" case with placeholders. Infer these placeholders and update the following keys:
{"prompt": <str>, "prompt_fill_blank": <bool>}

To update the prompt if it is "fill in the blank":
- If the prompt is not "fill in the blank", which should happy the majority of the time, you should by default copy the input prompt exactly and set "prompt_fill_blank" to False.
- Otherwise update the prompt and update the flag "prompt_fill_blank" to True.
- Often fill in the blank prompts include brackets of the form "[keyword]".
- Often you can infer the right keyword to replace the placeholder using the output images.
- Often you will generate multiple infilled prompts, because there are often multiple output images that represent different instantiations with different sets of keywords.
- Only fill in the blank only when it makes sense to do so, and when you are fairly confident about what the keyword should be. Otherwise, if highly uncertain, don't "fill in the blank".
- You should make a new sample for each new instantiation of the "fill in the blank". If there are multiple outputs that infill with different keywords, you should create multiple samples.

## Screenshots of Conversations
For special images that show a screenshot of a conversation with the image generator, mark their image id:
{"conversation": <id>}

To extract a "conversation":
- For each conversation, you should create a new sample that represents the task expressed in the conversation.
- If there exist multiple images showing screenshots of the same conversation, select the main one showing the most task information and omit the others.
- Combined related samples and their fields like "inputs", "outputs", "post_urls", "prompt" to minimize redundancy.
- A conversation is defined as a screenshot that shows a conversation (which may involve a prompt and image(s)) in OpenAI's ChatGPT window. 
- If the image shows any other platform, it is not a conversation.
- If the image generation task is not clear (e.g., the screenshot seems to be using ChatGPT's LLM rather than image generation capabilities, the screenshot is extremely low quality, the images are extremely small), it is also not a conversation.
- If the sample does not contain a conversation, set "conversation" to the empty string "".

prompt: <prompt>
images: <images>
images_to_posts: <images_to_posts>
extracted:
  
\end{lstlisting}
\end{tcolorbox}
\caption{Prompt for multimodal processing with GPT-4o~\citep{oai2024gpt4o}, described in~\autoref{subsec:processing_multimodal}.}
\label{fig:2_classify_images_fib_prompts}
\end{figure*}

\begin{figure*}
\centering
\scriptsize
\begin{tcolorbox}[colback=white, colframe=black!75!black, title=Instructions]
\begin{lstlisting}
You are an extractor of multimodal prompts for image generation.

Your job is to extract the text prompt and bounding boxes of individual images from screenshots of conversations with an image generator.
You will also be provided relevant text that may be helpful for determining the input and output images from the screenshot.

Please output only a valid JSON dictionary according to this schema:
```json {"prompt": <str>, "inputs": <list of bounding boxes>, "outputs": <list of bounding boxes>}```

To extract "prompt":
- If there is text, run OCR and extract the raw text input by the user exactly.
- The extracted text should produce a fully specified and coherent image generation task; ignore other irrelevant text.
- If there is no relevant text, output the empty string "".

To extract "inputs" and "outputs":
- Extract a list of bounding boxes for every individual image.
- Each bounding box should be formatted as [x1, y1, x2, y2]; (x1, y1) is the top-left and (x2, y2) is the bottom-right.
- Also sort bounding boxes as "inputs" vs. "outputs" of the image generator.
- The extracted images should produce a fully specified and coherent image generation task; ignore other irrelevant images.
- Each image should only appear once. Ignore exact duplicates.
- If there are no "inputs" output an empty list [].
- If there are no "outputs" output an empty list [].

relevant_text: <relevant_text>
images: <images>
extracted:
\end{lstlisting}
\end{tcolorbox}
\caption{Prompt for parsing screenshots of conversations with Qwen-2.5-VL~\citep{Qwen2.5-VL}, described in~\autoref{subsec:processing_multimodal}.}
\label{fig:3_extract_conversation_screenshot}
\end{figure*}

\clearpage
\section{Automatic Evaluation Metrics}
\label{app:eval_metrics}

\paragraph{Overall Metrics.} In~\autoref{fig:0903_autoeval} we depict the prompt used for VLM-as-a-judge in our overall benchmark evaluation. We follow the ``single answer grading setup'' of MT-Bench~\citep{zheng2023judging}, and convert absolute scores into pseudo pairwise comparisons, which can be used to compute the win rate.

\paragraph{Specialized Metrics.} We display our prompt for rating the accuracy of rendered text in~\autoref{fig:typographic_accuracy}, and classifying the applicability of each sample to each specialized metric in~\autoref{fig:metrics_classification}.

\begin{figure*}[h]
\centering
\scriptsize
\begin{tcolorbox}[colback=white, colframe=black!75!black, title=Instructions]
\begin{lstlisting}
Please act as an impartial judge and evaluate the quality of the image output produced by an image generation model in response to an input instruction (expressed via text and/or image(s)).

Begin your evaluation by forming your own expectation of what a good output should look like for the given prompt. Describe this briefly before judging the output.

Then compare the model's output with your expectation. Point out errors, inaccuracies, or failures to follow the instruction, and identify missing details that would make the output better satisfy the instruction.

Make sure to consider the following factors equally:
- **Prompt Following**: Does the output interpret the text correctly and execute the requested task accurately?
- **Reference Fidelity**: Does the output preserve key details from the input images when relevant?
- **Realism and Aesthetics**: Is the output photorealistic (e.g., accurate anatomy, no artifacts, etc.) and visually appealing (e.g., balanced colors, well-framed composition, etc.) when relevant?

After providing your explanation, please rate the output on a scale of 1 to 10 by strictly following this format: "[[rating]]", for example: "Rating: [[5]]".

<|The Start of Input Instruction|>
input_prompt: <input_prompt>
input_images: <input_images>
<|The End of Input Instruction|>

<|The Start of Model Output|>
output_image: <output_image>
<|The End of Model Output|>
\end{lstlisting}
\end{tcolorbox}
\caption{Prompt for automatic evaluation with GPT-4o~\citep{oai2024gpt4o}, Gemini 2.0~\citep{team2023gemini}, and Qwen2.5-VL-32B-Instruct~\citep{Qwen2.5-VL}, described in~\autoref{sec:model_comparison}. Our prompt closely follows the format from MT-Bench~\citep{zheng2023judging}, but adapted for rating image generation outputs.}
\label{fig:0903_autoeval}
\end{figure*}
\begin{figure*}[h]
\centering
\scriptsize
\begin{tcolorbox}[colback=white, colframe=black!75!black, title=Instructions]
\begin{lstlisting}
Check if all text in the image is accurate and readable.
For exact copy requests: spelling, punctuation, grammar match exactly, with no missing or extra characters, and text is not cropped.
For created text: content is coherent, relevant, and fits the available space and design.
Begin your evaluation by reading through the image and OCR the text.
Point out spelling errors, punctuation errors, grammar errors, and missing characters of the text.
Point out if the text is cropped.
Then, look at the image again and check if the text is coherent, relevant, and fits the available space and design.
Please rate the output on a scale of 1 to 10 by strictly following this format: "[[rating]]", for example: "Rating: [[5]]".

<|The Start of Input Instruction|>
input_prompt: <input_prompt>
<|The End of Input Instruction|>

<|The Start of Model Output|>
output_image: <output_image>
<|The End of Model Output|>
\end{lstlisting}
\end{tcolorbox}
\caption{Prompt for judging text rendering accuracy with GPT-4o~\citep{oai2024gpt4o}, described in~\autoref{subsec:specialized_metrics}.}
\label{fig:typographic_accuracy}
\end{figure*}
\begin{figure*}
\centering
\scriptsize
\begin{tcolorbox}[colback=white, colframe=black!75!black, title=Instructions]
\begin{lstlisting}
  For each metric in the provided list, decide if it is applicable for the given image generation instruction.
  The instruction can be defined with text and/or images. Some instructions may contain no input images.
  If is a metric is marked as applicable, it will be used as an axis to score and rank outputs for the given input.

  ## Metric List
  <metric_list>

  ## Output Format
  Respond only with a JSON dictionary containing all the metric names as keys, and the value 0 (is not applicable) or 1 (is applicable).
  Also include a short global rationale for your overall decision-making process.
  ```json
  {
    "<metric1>": <integer 0 or 1>,
    "<metric2>": <integer 0 or 1>,
    ...
    "rationale": "<a short rationale, 20 words or less>"
    "prompt": "<the input prompt repeated again>"
  }
  ```

  ## Your Turn
  task: <task>
  input_prompt: <input_prompt>
  input_images: <input_images>
  metrics:
\end{lstlisting}
The <metric\_list> is replaced with the defined metrics name, its description, and its applicability:
\begin{lstlisting}
Name: "Face Identity Preservation"
Description: 
  Check if the person's identity matches the reference or intended person, keeping facial structure and distinctive features the same.
  Examples to Penalize: Changes in hairstyle, beard length, scars, facial expression, accessories, etc. that do not match the reference.
Applicability: 
  This metric is often applicable, especially for tasks involving subject-driven generation.
  However, it is not applicable when:
  - The prompt does not explicitly or implicitly request face identity preservation.
  - No person's face is visible (because there are no people, or faces are occluded).
  - The task is stylization, where the creative freedom allows for many valid outputs and correctness is too subjective.
\end{lstlisting}
\begin{lstlisting}
Name: "No Color Shift"
Description: 
  Check if the overall color tone, brightness, and contrast match the reference or intended look.
  Examples to Penalize: Added yellow tint, overexposure, or darkening compared to the reference.
Applicability: 
  This metric is often applicable, especially for tasks like local editing.
  However, it is not applicable when:
  - The task is colorization or image-to-image translation, where color change is inherent to the task.
\end{lstlisting}
\begin{lstlisting}
Name: "Spatial Position Preservation"
Description: 
  Check if the structure and spatial layout of the reference are copied correctly, including positions, relative layout, and scale of key objects.
  Examples to Penalize: A dog is slightly moved from its original position during stylization; a table that was centered is shifted.
Applicability: 
  This metric is only applicable for tasks that involving image-to-image translation, stylization, or local editing that requires strict structure preservation.
  However, it is not applicable when:
  - The prompt does not expect the resulting image to be strictly preserving spatial structure with the reference image.
  - The prompt can allow some structure changes (eg, sketch-to-image, 2D-to-3D stylization)
\end{lstlisting}
\begin{lstlisting}
Name: "Text Rendering Accuracy"
Description: 
  Check if rendered text contains mistakes that hinder readability.
  Examples to Penalize: Characters are garbled; there are missing or extra characters; there is incorrect spelling or punctuation; there is incorrect grammar.
Applicability: 
  This metric is often applicable, but only when the prompt explicitly requests rendered text.
\end{lstlisting}
\end{tcolorbox}
\caption{
Prompt for classifying the set of applicable samples for each specialized metric with GPT-4o~\citep{oai2024gpt4o}, described in~\autoref{subsec:specialized_metrics}.
The number of applicable samples is as follows: Face Identity Preservation (244), No Color Shift (271), Spatial Position Preservation (180), Text Rendering (240).
}
\label{fig:metrics_classification}
\end{figure*}

\clearpage
\section{LLM Disclosure}
\label{app:llm_disclosure}
Some portions of this work were generated with the assistance of large language models (LLMs). Their primary role was to support editing, rephrasing, and formatting of existing text to improve clarity and readability. While human authors created and reviewed the core content, LLMs were used as a tool to streamline refinement and presentation. All factual information, analysis, and conclusions remain the responsibility of the authors, and every effort has been made to ensure accuracy and integrity.

\end{document}